\documentclass[runningheads, orivec]{llncs}
\usepackage[T1]{fontenc}

\usepackage{cite}
\usepackage{subcaption}
\usepackage[acronym,nomain,nonumberlist]{glossaries}

\usepackage{hyperref}
\hypersetup{
    colorlinks=true,
    linkcolor=blue,
    urlcolor=blue,
}
    
\usepackage{setspace}
\usepackage{siunitx}
\usepackage{verbatim}
\usepackage{lastpage}
\usepackage{amsfonts,amssymb, amsmath}
\usepackage{enumitem}
\usepackage{booktabs}
\usepackage{multirow}
\usepackage{tabularx}
\usepackage{wrapfig}

\usepackage[dvipsnames]{xcolor}
\usepackage{graphicx}

\usepackage{tikz}
\usetikzlibrary{arrows.meta}
\usetikzlibrary{shapes.geometric}
\usetikzlibrary{positioning}
\usetikzlibrary{angles}
\usetikzlibrary {matrix}
\usetikzlibrary{calc}
\usetikzlibrary{shadings}
\usetikzlibrary{decorations.pathreplacing}
\newcolumntype{C}[1]{>{\centering\let\newline\\\arraybackslash\hspace{0pt}}m{#1}}

\definecolor{b0cnn}{HTML}{0173B2}
\definecolor{b0trans}{HTML}{029E73}
\definecolor{vitcnn}{HTML}{D55E00}
\definecolor{vittrans}{HTML}{CC78BC}

\definecolor{b0cnn_edge}{HTML}{01507d}
\definecolor{b0trans_edge}{HTML}{016f50}
\definecolor{vitcnn_edge}{HTML}{954200}
\definecolor{vittrans_edge}{HTML}{8f5484}

\newcommand{\bs}[1]{\boldsymbol{#1}}
\newcommand{\narrowbold}[1]{%
  \textbf{\scalebox{0.9}[1]{#1}}%
}
\newcommand{\etal}{\textit{et al}.}

\newcommand{\std}[1]{{\,\scriptsize$\pm$#1}}

\begin{document}

\newacronym{cnn}{CNN}{Convolutional Neural Network}
\newacronym{knn}{$k$-NN}{$k$\mbox{-}Nearest Neighbors}
\newacronym{nn}{NN}{Nearest Neighbor}
\newacronym{lstm}{LSTM}{Long Short\mbox{-}Term Memory}
\newacronym{vit}{ViT}{Vision Transformer}
\newacronym{clip}{CLIP}{Contrastive Language\mbox{-}Image Pre\mbox{-}training}
\newacronym{clap}{CLAP}{Contrastive Language\mbox{-}Audio Pre\mbox{-}training}
\newacronym{ifcb}{IFCB}{Imaging FlowCytobot}

\newacronym{ifc}{IFC}{Imaging Flow Cytometry}
\newacronym{syke}{Syke}{Finnish Environment Institute}

\newacronym{sgd}{SGD}{Stochastic Gradient Descent}
\newacronym{ssl}{SSL}{Self\mbox{-}Supervised Learning}
\newacronym{dino}{DINO}{Self\mbox{-}DIstillation with NO Labels}

\newacronym{siglip}{SigLIP}{Sigmoid loss for
Language\mbox{-}Image Pre\mbox{-}training}

\title{Cross-modal learning for plankton recognition}

\author{
    Joona Kareinen\inst{1} \and
    Veikka Immonen\inst{1} \and
    Tuomas Eerola\inst{1} \and
    Lumi Haraguchi\inst{2} \and
    Lasse Lensu\inst{1} \and
    Kaisa Kraft \inst{2} \and
    Sanna Suikkanen \inst{2} \and
    Heikki K\"alvi\"ainen\inst{1,3}
}
\authorrunning{J. Kareinen et al.}

\institute{
Lappeenranta-Lahti University of Technology LUT, Lappeenranta, Finland \and
Finnish Environment Institute, Helsinki, Finland \and
Faculty of Information Technology, 
Brno University of Technology,
Czechia
}
\maketitle

\begin{abstract}
This paper considers self-supervised cross-modal coordination as a strategy enabling utilization of multiple modalities and large volumes of unlabeled plankton data to build models for plankton recognition. Automated imaging instruments facilitate the continuous collection of plankton image data on a large scale. Current methods for automatic plankton image recognition  rely primarily on supervised approaches, which require labeled training sets that are labor-intensive to collect. On the other hand, some modern plankton imaging instruments complement image information with optical measurement data, such as scatter and fluorescence profiles, which currently are not widely utilized in plankton recognition. In this work, we explore the possibility of using such measurement data to guide the learning process without requiring manual labeling. Inspired by the concepts behind \gls{clip}, we train encoders for both modalities using only binary supervisory information indicating whether a given image and profile originate from the same particle or from different particles. For plankton recognition, we employ a small labeled gallery of known plankton species combined with a $k$-NN classifier. This approach yields a recognition model that is inherently multimodal, i.e., capable of utilizing information extracted from both image and profile data. We demonstrate that the proposed method achieves high recognition accuracy while requiring only a minimal number of labeled images. Furthermore, we show that the approach outperforms an image-only self-supervised baseline. Code available at \url{https://github.com/Jookare/cross-modal-plankton}.
\keywords{plankton recognition  \and multimodal learning \and coordinated representation}
\end{abstract}

\section{Introduction}
Plankton play a fundamental role in sustaining life on Earth, contributing ca. 50\% of global oxygen and ca. 40\% of carbon fixation, forming the base of aquatic food webs, and serving as sensitive indicators of environmental change~\cite{falkowski1994role, field1998primary}. As climate change and pollution increasingly impact aquatic ecosystems, monitoring plankton communities can provide early warnings of ecological disruption. Yet, despite their importance, plankton communities remain poorly understood due to their rapid turnover and sensitivity to environmental changes. 

Over the past decades, technological advancements have led to the development of several automated plankton imaging instruments that combine microscopy with digital cameras to produce high-quality images at relatively low operational costs. These devices enable large-scale data collection, generating millions of plankton images across different aquatic environments~\cite{orenstein2015whoi, kraft2022ifcb, gallot2025best}. However, manual annotation at such a scale is infeasible, as expert annotation of plankton images is both expensive and time-consuming. To address this limitation, various computer vision-based plankton recognition methods have been developed~\cite{eerola2024survey}.

Most existing plankton recognition methods rely on a single modality of data, typically images. While image-based models have achieved accurate results, they are inherently limited by image quality, subtle inter-species differences, and incomplete class coverage in labeled datasets~\cite{eerola2024survey}. Some plankton imaging instruments, however, provide additional data beyond images, such as optical measurements (e.g., scatter and fluorescence profiles), or are associated with contextual metadata (e.g., sampling location, water temperature, and time)~\cite{dubelaar1999cytosense}. Some instruments, like pulse-shape recording and imaging flow cytometer CytoSense~\cite{dubelaar1999cytosense, gallot2025best}, rely on the optical measurements as their primary modality, which is recorded for each particle, and the images are taken as supplementary to help connect the optical measurements with taxonomical units. Although these additional or auxiliary modalities contain complementary information about both individual particles and entire samples, they are underutilized in existing recognition approaches. Data composed primarily of optical information is also more challenging for existing models to recognize. The existence of this additional data creates an opportunity for multimodal learning, where recognition models leverage multiple data types simultaneously or interchangeably, potentially improving both accuracy and robustness.

Perhaps even more notably, the auxiliary data can be used to guide self-supervised learning, thereby reducing the need for labor-intensive expert labeling. Cross-modal learning techniques, such as \acrfull{clip}~\cite{radford2021clip} align representations across modalities, enabling classification or retrieval without class labels. This characteristic makes the use of multiple modalities particularly compelling: each particle is described by complementary signals that provide a richer representation than image data alone. At the same time, it allows to take advantage of the large volumes of unlabeled data produced by modern imaging instruments for model training.%without the need for prior labeling.

\begin{figure}[tb]
    \centering
    \includegraphics[width=\linewidth]{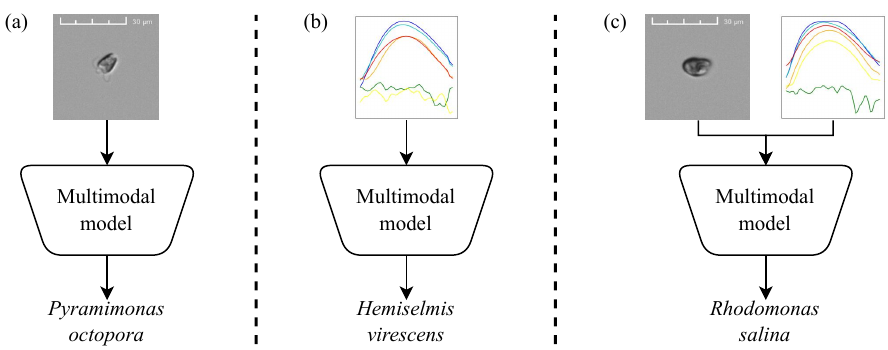}
    \caption{Overview of the proposed multimodal plankton recognition model for CytoSense capable of classifying plankton species using (a) only bright-field images, (b) only optical profiles, or (c) both at the same time.}
    \label{fig:multimodal_model}
\end{figure}

In this paper, we study cross-modal learning for plankton recognition, focusing specifically on data collected using CytoSense~\cite{dubelaar1999cytosense, gallot2025best}, which includes both bright-field images and optical profiles of individual particles. 
We aim to address three key research questions: (1) Is it possible to obtain image and profile embeddings suitable for plankton recognition using cross-modal learning without relying on class labels? (2) Can plankton recognition be improved by incorporating an additional modality? (3) How well does the framework perform with only the non-image data? To answer these questions, and inspired by \gls{clip}, we propose a contrastive pre-training approach that aligns paired image--profile samples in a shared embedding space. For plankton recognition, we introduce a method that uses small, labeled gallery of known plankton species and performs nearest neighbor search in the embedding space. This recognition method allows  either modality to be used independently or both modalities to be combined for improved performance (see Fig.~\ref{fig:multimodal_model}).

In the experimental section, we demonstrate that the proposed method not only achieves high plankton recognition accuracy when training and test data come from the same dataset (in-domain) but also generalizes well to different datasets with varying class distributions (cross-domain). Furthermore, we show that the method outperforms the image-only self-supervised learning baseline DINO~\cite{caron2021emerging}.

The main contributions of this study are as follows: (1) a multimodal plankton recognition framework for CytoSense data that combines bright-field images and optical profiles, (2) a new publicly available multimodal plankton dataset, SYKE-plankton\_CytoSense\_2025~\cite{cytosensedata}, and (3) a label-free contrastive cross-modal learning strategy, accompanied with an analysis of how different modalities contribute to recognition performance.

\section{Related works}
\subsection{Plankton recognition}

As automated plankton imaging instruments now enable large-scale and even real-time monitoring of plankton communities, efficient automatic recognition methods are essential. Existing approaches range from early feature-engineering methods~\cite{sosik2007automated} to modern deep learning models~\cite{eerola2024survey}. Current state-of-the-art techniques typically employ either \glspl{cnn}~\cite{kraft2022ifcb} or transformer-based architectures such as \glspl{vit}~\cite{kyathanahally2022ensembles}, often combined with self-supervised or transfer learning strategies~\cite{kyathanahally2022ensembles, maracani2023indomain, kareinen2025selfsupervised}. Research on plankton recognition has mainly focused on developing models for image data, with studies related to recognition of the pulse shape profiles being scarce~\cite{fuchs2022automatic}

Most existing work on plankton recognition has focused on developing models for one or a few specific datasets using labeled training sets. Consequently, numerous models have been proposed that perform well on a single imaging instrument and a fixed set of species but fail to generalize to other datasets. Creating new manually annotated image datasets for instrument deployments in new locations with varying species composition is impractical, highlighting the need for more generalizable methods. Recent efforts have addressed this shortcoming through open-set recognition~\cite{badreldeen2022open,kareinen2025open,YangZhenyu2022Clir} and domain adaptation~\cite{batrakhanov2024daplankton, chen2025producing}. However, developing robust and generalizable models that minimize manual expert labeling remains essential for real-world plankton monitoring.

\subsection{Multimodal learning}

Multimodal machine learning utilizes a range of methods to jointly process and integrate information from multiple data sources, or modalities, (e.g., images, text, audio, sensor data) to produce context-aware representations for tasks like visual question answering and cross-modal retrieval~\cite{liang2024surveymultimodal}. A common approach is representation coordination, which aligns modality-specific representations in a shared embedding space to enforce semantic consistency across modalities while preserving their individual characteristics. This is typically achieved by using separate encoders for each modality and applying alignment constraints only at the embedding level~\cite{liang2024surveymultimodal}. 

Contrastive learning has become the dominant paradigm for representation coordination, starting from image--sentence alignment~\cite{kiros2014unifying} and culminating in large-scale models such as \gls{clip}~\cite{radford2021clip}. These models maximize similarity between matching pairs (e.g., an image and its caption) and minimize it for non-matching pairs, enabling strong zero-shot generalization. Subsequent models, including \gls{siglip}~\cite{zhai2023siglip} and SigLIP 2~\cite{tschannen2025siglip}, further improved optimization and scalability. Beyond vision--language, similar principles extend to other modality pairs, such as audio--text in \gls{clap}~\cite{elizalde2023clap}, demonstrating the versatility of contrastive multimodal learning.

In the biological domain, Bio\gls{clip}~\cite{stevens2024bioclip} and Bio\gls{clip} 2~\cite{gu2025bioclip2} adapt the \gls{clip} framework for biological imaging tasks. Trained on millions of organism images paired with taxonomic labels, these models enable zero-shot classification across diverse biological datasets. Despite its success in broader biological contexts, Bio\glspl{clip} have shown limited effectiveness in direct plankton recognition~\cite{stevens2024bioclip, gu2025bioclip2}.

Although multimodal learning has not been widely studied in plankton recognition, some studies have combined image data with supplementary metadata. These works typically use simple fusion strategies such as feature or input concatenation rather than coordinated representations. For example, Ellen \etal~\cite{ellen2019improving} combined image features with geotemporal and hydrographic metadata and Bures \etal~\cite{bures2021plankton} incorporated image size and geotemporal metadata. These works represent initial attempts to leverage multimodal information for plankton recognition, although they rely solely on metadata rather than true multi-sensor modalities, which is the focus of this study.

\section{Methods}

\begin{figure}[tb]
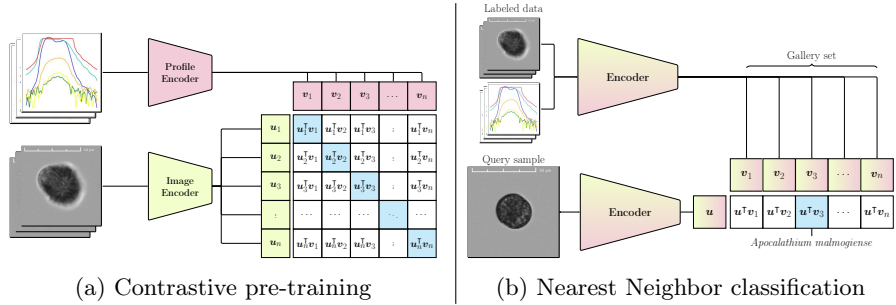

    \begin{subfigure}[b]{0.47\textwidth}
         \centering
        \resizebox{\textwidth}{!}{
            \input{images/pretrainig}%
        }
         \caption{Contrastive pre-training}
    \end{subfigure}
    \hspace{0.1cm}\vrule\hspace{0cm}
    \begin{subfigure}[b]{0.47\textwidth}
         \centering
        \resizebox{\textwidth}{!}{
            \input{images/zeroshot}%
        }
         \caption{Nearest Neighbor classification}
    \end{subfigure}
    \caption{Overview of the proposed multimodal recognition framework. (a) During pre-training, paired image--profile samples are aligned in a shared embedding space. (b) At inference, classification is performed using nearest-neighbor search in the embedding space.}
    \label{fig:method}
\end{figure}

The proposed method adapts contrastive multimodal learning to plankton recognition. The pipeline consists of two stages: (1) contrastive pre-training to learn a shared multimodal embedding space, and (2) metric-based classification using nearest neighbors. The overall methodology is illustrated in Fig.~\ref{fig:method}.

\subsection{Contrastive Image--Profile Pre-training}
The pre-training stage adapts the contrastive learning framework of \gls{clip}~\cite{radford2021clip} by aligning plankton images with their corresponding optical profiles rather than text. In this setup, each image--profile pair represents a single plankton particle, and the model learns a shared embedding space where paired samples are close and unpaired samples are far apart. Unlike \gls{clip}, which leverages language to structure embeddings for open-vocabulary tasks, our approach uses physical measurements to capture domain-specific properties of plankton data.

Formally, let $\mathcal{B} := \{(\bs{x}_i, \bs{x}_p)\}_{i=1}^n$ be a batch of $n$ multimodal samples, where $\bs{x}_i$ is an image and $\bs{x}_p$ the corresponding profile. Two modality-specific encoders, $f_i$ and $f_p$, produce intermediate representations:
\[
\bs{r}_i = f_i(\bs{x}_i), \quad \bs{r}_p = f_p(\bs{x}_p),
\]
where $\bs{r}_i \in \mathbb{R}^{u}$ and $\bs{r}_p \in \mathbb{R}^{v}$. These representations are projected into a shared $d$-dimensional space via linear layers $L_i$ and $L_p$:
\[
\bs{e}_i = L_i(\bs{r}_i), \quad \bs{e}_p = L_p(\bs{r}_p), \quad \bs{e}_i, \bs{e}_p \in \mathbb{R}^d.
\]

After $\ell_2$ normalization, similarity between two embeddings is measured using cosine similarity:
\[
S(\bs{e}_i, \bs{e}_p) = \frac{\bs{e}_i^\top \bs{e}_p}{\|\bs{e}_i\|\|\bs{e}_p\|}.
\]

\subsubsection{InfoNCE loss.}
We adopt the InfoNCE contrastive loss~\cite{oord2018representation}, which has been widely used in multimodal representation learning. 
The loss encourages matched image--profile pairs to have higher similarity than mismatched pairs in the embedding space. The loss is computed using symmetric cross-entropy loss for both modalities. For image-to-profile alignment:
\[
\mathcal{L}_{i \rightarrow p} = -\frac{1}{n} \sum_{k=1}^{n} \log \frac{\exp( \tau S(\bs{e}_i^k, \bs{e}_p^k))}{\sum_{j=1}^{n} \exp(\tau S(\bs{e}_i^k, \bs{e}_p^j))},
\]
where $\tau$ is a learnable temperature parameter. The profile-to-image  alignment loss $\mathcal{L}_{p \rightarrow i}$ is computed in the same way, and the final loss is the average of the two.

\subsubsection{Sigmoid loss.}
Additionally, we adopt the sigmoid loss introduced in \gls{siglip}~\cite{zhai2023siglip}, which replaces the softmax normalization with independent binary classification for all pairs.  The objective is a sigmoid-based binary cross-entropy:
\[
\mathcal{L}_\text{Sigmoid} = -\frac{1}{n}\sum_{i=1}^n \sum_{j=1}^n \log \frac{1}{1 + \exp(-z_{i,j}(\tau S(\bs{e}_i, \bs{e}_p) + b))},
\]
where $z_{i,j} = 1$ when $i=j$ (positive pair), and $-1$ otherwise. Here $\tau$ is a learnable temperature parameter, and $b$ is a learnable bias term that improves stability during early training.

\subsection{Nearest Neighbor classification}
After pre-training, the learned multimodal embedding space is reused for classification. In contrast to the original \gls{clip}, which compares image embeddings against a fixed set of text embeddings, here we construct a gallery of labeled multimodal embeddings.

Formally, a gallery set is generated by encoding labeled plankton samples from all classes and modalities with the pretrained encoders. Each gallery embedding serves as a reference to a known class. Given a new test sample $\bs{v}$, its embeddings are compared to the gallery embeddings $\bs{u_i}$ using cosine distance
\[
D(\bs{u_i}, \bs{v}) = 1 - S(\bs{u_i}, \bs{v}),
\]
and classification is based on the nearest neighbors. To improve robustness, $k$-nearest neighbor voting is used instead of a single nearest neighbor.

The framework naturally supports multimodal test samples. Each modality is encoded separately, and $k$ neighbors are retrieved for each embedding. The neighbors are then combined, and the final prediction is done based on the majority vote. Compared to the text-based lookup in \gls{clip}, this design places no restrictions on the number of reference samples per class, and benefits directly from a larger gallery set.

\section{Experiments}
\subsection{Data}
\begin{figure}[htb]
    \begin{subfigure}[b]{0.32\textwidth}
         \centering
         \includegraphics[width=\textwidth]{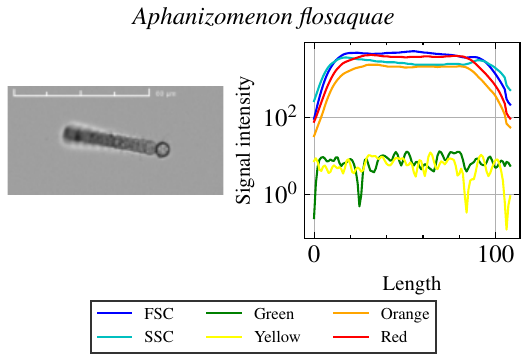}
         \caption{}
    \end{subfigure}
    \hfill
    \begin{subfigure}[b]{0.32\textwidth}
         \centering
         \includegraphics[width=\textwidth]{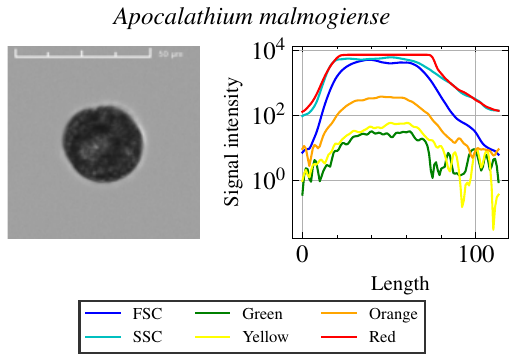}
         \caption{}
    \end{subfigure}
    \hfill
    \begin{subfigure}[b]{0.32\textwidth}
         \centering
         \includegraphics[width=\textwidth]{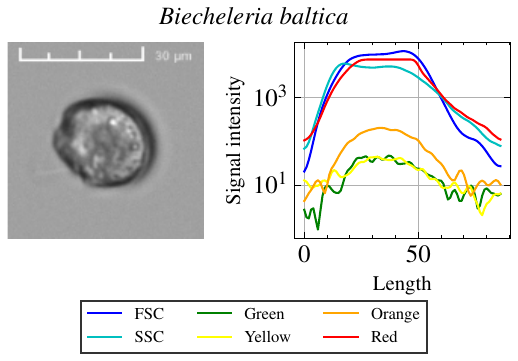}
         \caption{}
    \end{subfigure}
    \caption{Example image and profile data}
    \label{fig:cyto}
\end{figure}

Three multimodal plankton datasets acquired with the CytoSense (CS) instrument are utilized~\cite{dubelaar1999cytosense, gallot2025best}. CytoSense is a flow cytometer designed for multimodal plankton analysis. The device provides optical profiles for each particle with a complementary set of bright-field images. For this study, we used a dataset with each sample (data point pair) consisting of both:
\begin{enumerate}
    \item \textbf{Bright-field image:} a gray-scaled image, analyzed on a size range of \SI{0.5}{}-\SI{800}{\micro\metre}, covering most phytoplankton particles.
    \item \textbf{Optical profile}: a set of six optical signals measured as each particle passes through the laser. Signals are excited by a blue (\SI{488}{\nano\metre}) and an orange laser (\SI{596}{\nano\metre}) and detected by sensors for Forward Scatter (FSC), Sideward Scatter (SSC), and fluorescence in the green (\SI{502}{}–\SI{538}{\nano\metre}), yellow (\SI{553}{}–\SI{577}{\nano\metre}), orange (\SI{604}{}–\SI{644}{\nano\metre}), and red (\SI{668}{}-\SI{726}{\nano\metre}) channels. 
\end{enumerate}
The six optical signals can be represented as a sequence of $T$ optical feature vectors
\[
\bs{P} = \begin{bmatrix} \bs{p}_1 & \dots & \bs{p}_T \end{bmatrix}, 
\quad
\bs{p}_t = \begin{bmatrix} \mathrm{FSC}(t) \\ \mathrm{SSC}(t) \\ \mathrm{Green}(t) \\ \mathrm{Yellow}(t) \\ \mathrm{Orange}(t) \\ \mathrm{Red}(t) \end{bmatrix},
\]
where $t$ denotes the position along the particle. In the image $\bs{I}\in \mathbb{R}^{H\times W}$, the length axis of the plankton is oriented horizontally. Example image--profile pairs are shown in Fig.~\ref{fig:cyto}.
 
In total, we utilize three multimodal datasets, namely the LAB, SEA, and UTO datasets, and publish them as a collection under SYKE-Plankton\_CytoSense\_2025 \cite{cytosensedata}. Two datasets are labeled and one unlabeled. An overview of the datasets is presented in Table~\ref{tab:dataset_summary}.

\begin{table}[htb]  
    \centering
    \caption{Summary of the datasets used in the experiments.
    }
    \resizebox{\textwidth}{!}{
    \begin{tabular}{p{2cm} C{3cm} C{3cm} C{3cm} C{3cm}}
        \toprule
        \textbf{Dataset} & Instrument & Number of classes & Samples per class & Total \\ 
        \midrule
        LAB & CS-2019-97 & 24 & 543 -- 1,285 & 20,050  \\
        SEA & CS-2019-97 & 38 & 1 -- 2,594 & 9,353 \\
        UTO & CS-2021-103 & unknown & unknown & 32,930    \\
        \bottomrule
    \end{tabular}%
    }
    \label{tab:dataset_summary}
\end{table}

The \textbf{LAB} dataset consists of 20,050 samples from 24 monospecific phytoplankton cultures obtained from the FINMARI Culture Collection of Syke Marine Research Laboratory and Tvärminne Zoological Station in March 2022. The dataset shares the same source material as DAPlankton$_{\textrm{LAB}}$~\cite{batrakhanov2024daplankton} but includes a larger set of classes, as it is not restricted to species present across multiple imaging instruments.

The \textbf{SEA} dataset consists of 9,353 samples across 38 classes collected \textit{in situ} in the Baltic Sea in July 2020. Similarly to the \textbf{LAB} dataset, it shares the source material with DAPlankton$_{\textrm{SEA}}$~\cite{batrakhanov2024daplankton}. The dataset is heavily imbalanced, with the largest class accounting for over 25\% of all samples. Compared to DAPlankton$_{\textrm{SEA}}$, the SEA dataset includes additional classes but fewer total samples, as some images were missing the profile information.

The \textbf{UTO} dataset consists of 32,930 CytoSense samples collected \textit{in situ} in the Baltic Sea between March 2024 and October 2024. The instrument was sampling hourly in an automated mode from the flow through system at the Utö Marine and Atmospheric Research Station (for details see \cite{kraft2025monitoring}). The dataset contains raw, unfiltered data used exclusively for self-supervised pre-training.

\subsection{Data pre-processing and augmentation}
Preprocessing and augmentation were designed to maintain cross-modal consistency between images and optical profiles. For images, the scale bar was cropped, the remaining region padded to a square using edge padding to preserve aspect ratio, and the square resized to $236\times236$ pixels. Images were converted to grayscale and randomly cropped to $224\times224$ with a scale sampled from $(0.8,1.0)$. Random vertical flips and color jitter, altering brightness and contrast within $(0.8,1.2)$, were applied, followed by normalization to the range $(-1,1)$.

\begin{figure}[tb]
    \begin{subfigure}[b]{0.49\textwidth}
         \centering
         \begin{tikzpicture}
            \node[anchor=south west,inner sep=0] (img)
                {\includegraphics[width=\textwidth]{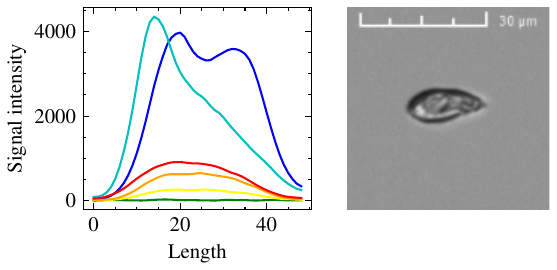}};
            % Overlay label
            \node[anchor=south, fill=white, opacity=0.8, text opacity=1]
                at ($(img.south east)!0.5!(img.south west)+(0,-0.2)$) {(a)};
        \end{tikzpicture}
    \end{subfigure}
    \hfill
    \begin{subfigure}[b]{0.49\textwidth}
         \centering
         \begin{tikzpicture}
            \node[anchor=south west,inner sep=0] (img)
                {\includegraphics[width=\textwidth]{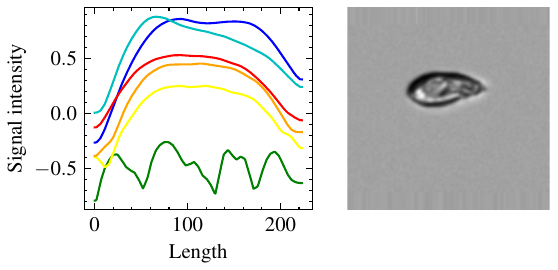}};
            % Overlay label
            \node[anchor=south, fill=white, opacity=0.8, text opacity=1]
                at ($(img.south east)!0.5!(img.south west)+(0,-0.2)$) {(b)};
        \end{tikzpicture}
    \end{subfigure}
    \caption{Pre-processing and data augmentations: (a) original profile and image, (b) same pair pre-processed}
    \label{fig:augmentation}
\end{figure}

For optical profiles, each signal was resampled to a fixed length of $224$, transformed with logarithmic compression $\log(1+x)$ to reduce magnitude variation while preserving relative shape, and normalized to $(-1,1)$. Two stochastic augmentations were applied: random amplitude scaling, which perturbs the optical signal intensity, and random band dropping, which zeroes out some of the signals. Finally, random horizontal flipping was applied synchronously to both modalities. An example of original and pre-processed samples is shown in Fig.~\ref{fig:augmentation}.

\subsection{Experimental Setup}
We systematically evaluated our proposed pre-training strategy by combining different image encoders, profile encoders, and loss functions to assess in-domain and cross-domain performance. The goal was to determine how architectural and loss-function choices influence multimodal representation quality. Specifically, we compared three image encoders: EfficientNet B0~\cite{tan2019efficient}, ConvNeXt Femto~\cite{liu2022convnet}, and \gls{vit}-Tiny~\cite{dosovitskiy2020vit}, representing convolutional, modern convolutional, and transformer-based architectures, respectively. These variants were chosen to maintain similar model sizes because our training datasets were relatively small, making lightweight architectures more suitable ($5-7$M parameters). For the optical profile modality, we evaluated three sequence encoders: 1D \gls{cnn}, bidirectional \gls{lstm}~\cite{hochreiter1997lstm}, and Transformer~\cite{vaswani2017transformer} (architectural details shown in Appendix \ref{appendix:A1}). Lastly, two contrastive loss formulations were tested: InfoNCE~\cite{oord2018representation} and Sigmoid loss~\cite{zhai2023siglip}.

In addition to the learned representations from each encoder, we appended the original image size (prior to cropping and padding) and the raw scattering profile length (prior to resampling) to the respective modality feature vectors before the projection heads. These scalar features were divided by the final target size ($224$), producing normalized values that encode the relative scale of each sample with respect to the resized input.

All models were trained using \gls{sgd} with Nesterov momentum~\cite{sutskever2013nesterov} of $0.9$ and weight decay of $0.001$. The learning rate was set to $0.005$ for InfoNCE and $0.002$ for Sigmoid loss and scheduled using a cosine annealing with $5$ epoch linear warm up. Training was performed for up to $100$ epochs, unless otherwise stated, using a batch size of $256$, with early stopping applied if the validation loss did not improve for $30$ consecutive epochs. $512$ was used as the common embedding space dimension. 

Throughout the experiments, we use the notation: [training set] $\rightarrow$ [test set]. For example, LAB $\rightarrow$ SEA indicates that the model was trained on the LAB dataset and tested on SEA. We use similar notation to denote modalities in the gallery and query sets (e.g., I+P $\rightarrow$ I means the gallery contains both modalities, while the query set contains only images). In all results, we used a gallery set of $16$ samples, $k$-NN $k$ value of $3$, and randomly resampled the gallery set $10$ times to mitigate sampling bias.

\subsection{Results}
\subsubsection{In-domain Evaluation.}
The first experiment, evaluated in-domain performance on the LAB dataset using 5-fold cross-validation with stratified splits: 80\% training, 5\% validation, and 15\% testing. Table~\ref{tab:lab_lab} presents the average accuracies and standard deviations for different encoder combinations and loss functions. Full results with varying modalities in gallery and test sets are given in Appendix~\ref{appendix:A2}.
Notably, the InfoNCE loss consistently outperforms the Sigmoid loss across all configurations, indicating its superior ability to align image and optical profile modalities in a shared embedding space, especially in this low-data regime. Of the profile encoders, CNN achieved the highest accuracy, followed by Transformer and LSTM. In the later experiments, we focused the evaluation on the CNN and Transformer profile encoders and exclusively used InfoNCE loss, as LSTM and the Sigmoid loss consistently yielded lower performance.

\begin{table}[b]
    \centering
    \caption{In-domain average accuracy and standard deviation (\%) for LAB $\rightarrow$ LAB with I+P $\rightarrow$ I+P}
    \begin{tabular}{l p{3cm} C{2.5cm} C{2.5cm} C{2.5cm}}
        \toprule
        Loss & Model & CNN & LSTM & Transformer \\
        \midrule
        \multirow{3}{1.5cm}{InfoNCE}
        & EfficientNet B0 & 95.06\std{0.58} & 89.59\std{0.67} & 92.11\std{0.98} \\
        & ConvNeXt Femto & 94.81\std{0.90} & 90.10\std{0.67} & 89.85\std{3.84} \\
        & ViT Tiny & \narrowbold{96.01\std{0.56}} & 90.39\std{0.66} & 91.80\std{0.76} \\
    \end{tabular}
    \begin{tabular}{l p{3cm} C{2.5cm} C{2.5cm} C{2.5cm}}
        \midrule
        \multirow{3}{1.5cm}{Sigmoid}
        & EfficientNet B0 & 91.85\std{0.65} & 84.13\std{0.77} & 85.98\std{1.28} \\
        & ConvNeXt Femto & 89.93\std{1.81} & 85.82\std{1.50} & 86.56\std{2.54} \\
        & ViT Tiny & 94.89\std{0.43} & 87.45\std{1.15} & 90.24\std{1.27} \\
        \bottomrule
    \end{tabular}
    \label{tab:lab_lab}
\end{table}

Table~\ref{tab:lab_lab_full_vit_tiny_cnn} provides the complete breakdown for the best-performing configuration identified above: ViT-Tiny as the image encoder, CNN as the profile encoder, and InfoNCE as the loss function. The highest accuracy was achieved when both images and profiles are present in both the gallery and test sets (I+P $\rightarrow$ I+P), confirming the benefit of multimodal recognition. Additionally, using only profiles in the gallery set yields better performance than using only images. This advantage is, however, largely due to the LAB dataset containing species with distinct optical characteristics. Somewhat surprisingly, recognition accuracy remains high even when the gallery and test sets use different modalities (e.g., I $\rightarrow$ P), suggesting effective modality alignment. The full results for all configurations are presented in the supplementary material. 

\begin{table}[tb]
    \centering
    \caption{Average accuracy and standard deviation for the best configuration (ViT-Tiny, CNN, InfoNCE) on LAB $\rightarrow$ LAB}
    \begin{tabular}{l C{2.5cm} C{2.5cm} C{2.5cm}}
        \toprule
        Gallery | Test & I & P & I+P \\
        \midrule
        I    & 92.73\std{0.59} & 93.03\std{0.69} & 94.24\std{0.59} \\
        P    & 91.56\std{0.67} & 94.51\std{0.83} & 95.42\std{0.73} \\
        I+P  & 92.79\std{0.57} & 94.55\std{0.82} & \narrowbold{96.01\std{0.56}} \\
        \bottomrule
    \end{tabular}
    \label{tab:lab_lab_full_vit_tiny_cnn}
\end{table}

\noindent
\textbf{Cross-domain Evaluation.}
Table~\ref{tab:cross_domain} presents the cross-domain performance of the image and optical profile encoder pairs trained with the InfoNCE loss (Full results given in Appendix~\ref{appendix:A3}). The most notable trend is that models pretrained on the UTO dataset generalize better to SEA than those trained on LAB. This result suggests that the stable growing conditions of the cultures used in the LAB dataset yield less variability (e.g., due to physiology and diversity), thus representing only a fraction of the heterogeneity found in the wild, and limiting the generalization of the LAB dataset to real-world plankton distributions. In contrast, the results for the UTO dataset demonstrates that unlabeled data can produce strong multimodal embeddings without manual annotations. As regards architectural choices, differences between the image encoders remain small, while for profile encoding CNN tends to outperform Transformer when applied to the SEA dataset.

\begin{table}[tb]
    \centering
    \caption{Cross-domain average accuracy and standard deviation (\%) for different image and profile encoder combinations with I+P $\rightarrow$ I+P}
    \resizebox{\textwidth}{!}{
    \begin{tabular}{p{3cm} *{6}{C{2.1cm}}}
        \toprule
        \multirow{2}{*}{Model} &
        \multicolumn{2}{c}{LAB $\rightarrow$ SEA} &
        \multicolumn{2}{c}{UTO $\rightarrow$ LAB} &
        \multicolumn{2}{c}{UTO $\rightarrow$ SEA} \\
        \cmidrule(lr){2-3} \cmidrule(lr){4-5} \cmidrule(lr){6-7}
        & CNN & Transf. & CNN & Transf. & CNN & Transf. \\
        \midrule
        EfficientNet B0 & 67.57\std{2.23} & 66.87\std{2.07} & 84.22\std{0.99} & \narrowbold{88.35\std{0.70}} & \narrowbold{72.86\std{1.85}} &  69.04\std{2.01} \\
        ConvNeXt Femto  & \narrowbold{68.05\std{2.06}} & 64.49\std{3.42} & 83.73\std{0.79} & 88.13\std{0.74} & 70.63\std{1.77} & 68.90\std{2.27} \\
        ViT Tiny        & 66.75\std{2.21} & 66.42\std{1.90} & 86.81\std{0.83} & 88.30\std{0.71} & 71.75\std{1.48} & 69.14\std{1.75} \\
        \bottomrule
    \end{tabular}}
    \label{tab:cross_domain}
    
\end{table}

\subsubsection{Combined Training.}
Table~\ref{tab:uto+lab_sea} shows the performance obtained by training the encoders with LAB and UTO datasets and evaluating them on the SEA dataset, including results for both the original model sizes and their larger variants ($18$-$28$\,M parameters): EfficientNet-B4, ConvNeXt-Tiny, and ViT-Small. Full results are shown in Appendix~\ref{appendix:A4}. Larger models were included to assess whether increased capacity improves performance when the amount of training data is larger. For larger models, the profile encoders were also scaled to keep the relative size of the encoders same. The results indicate that the CNN-based profile encoder performs worse than UTO-only training, whereas the Transformer-based encoder shows improvement. For larger models, we observe the opposite trend, suggesting that the available training data may be insufficient for the larger Transformer-based profile encoder.

\begin{table}[tb]
    \centering
    \caption{Average accuracy and standard deviation (\%) for combined training, UTO + LAB $\rightarrow$ SEA  with I+P $\rightarrow$ I+P}
    \begin{tabular}{p{3cm} C{2.5cm} C{2.5cm}}
        \toprule
         Model & CNN & Transformer \\
        \midrule
        EfficientNet B0 & 71.05\std{2.87} & 72.13\std{2.03} \\
        ConvNeXt Femto & 71.77\std{2.42} & 71.67\std{1.53} \\
        ViT Tiny & 71.37\std{2.17} & \narrowbold{72.73\std{1.20}} \\
    \end{tabular}
    \begin{tabular}{p{3cm} C{2.5cm} C{2.5cm}}
        \midrule
        EfficientNet B4 & \narrowbold{74.94\std{2.33}} & 71.78\std{2.99} \\
        ConvNeXt Tiny & 73.77\std{2.24} & 71.74\std{2.39} \\
        ViT Small & 72.75\std{2.04} & 71.04\std{1.40} \\
        \bottomrule
    \end{tabular}
    \label{tab:uto+lab_sea}
\end{table}

\subsubsection{Effect of Gallery Set Size.}

\begin{figure}[tb]
    \centering
    \hspace{1cm}
    \begin{subfigure}[b]{0.35\textwidth}
        \centering
        \includegraphics[width=\textwidth]{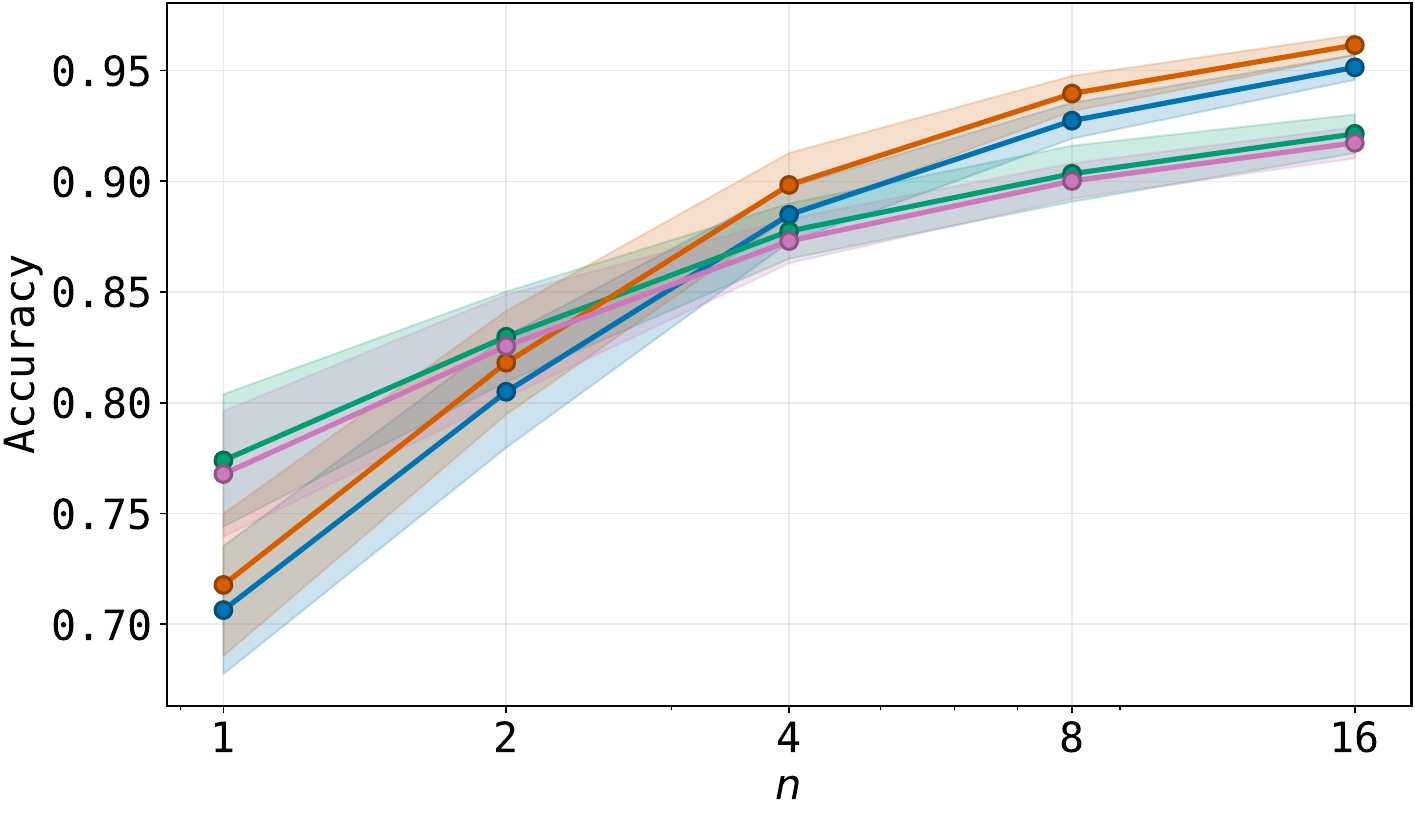}
        \caption{LAB $\rightarrow$ LAB}
    \end{subfigure}
    \hfill
    \begin{subfigure}[b]{0.35\textwidth}
        \centering
        \includegraphics[width=\textwidth]{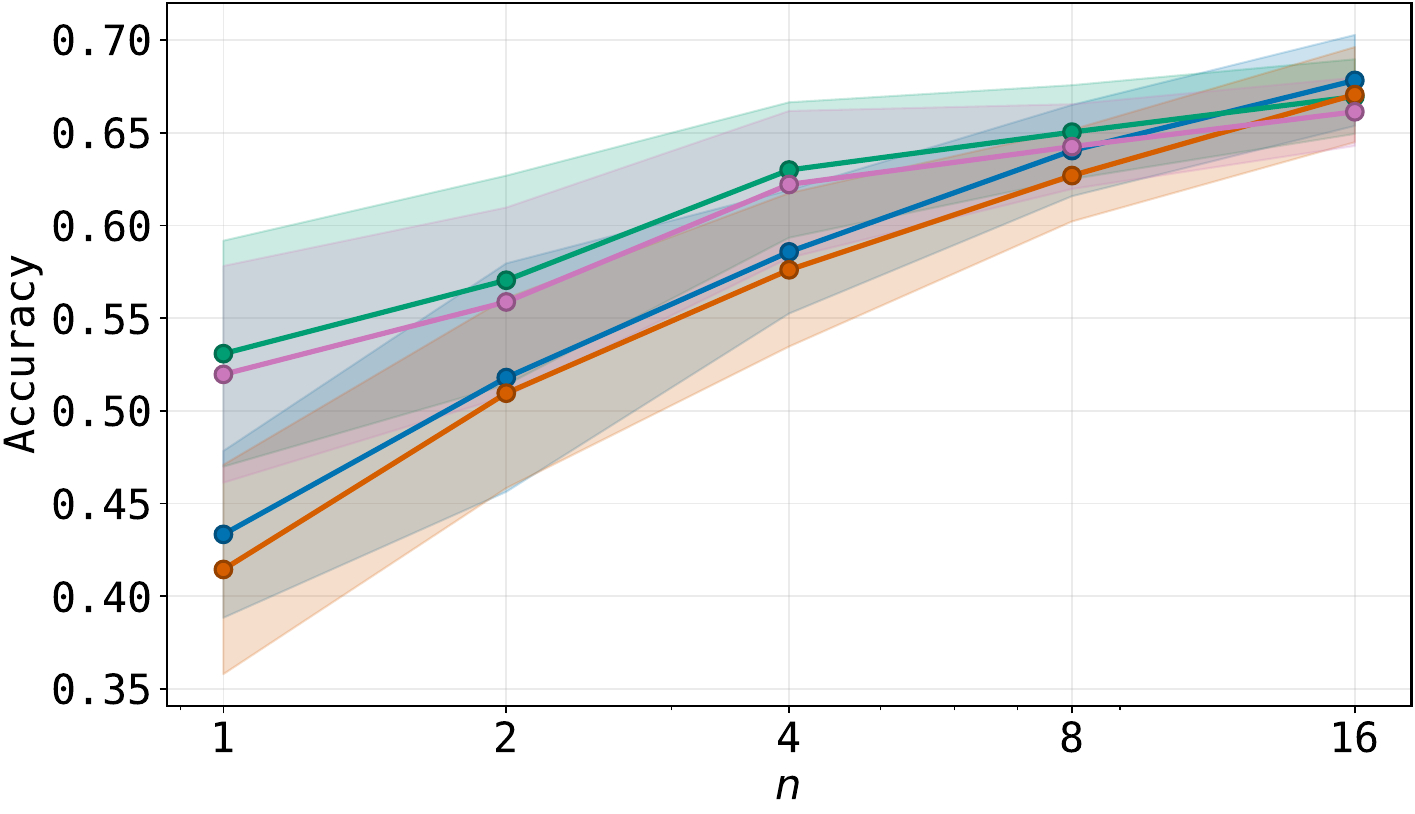}
        \caption{LAB $\rightarrow$ SEA}
    \end{subfigure}
    \hspace{1cm}
     \vskip\baselineskip
    \hspace{1cm}
    \begin{subfigure}[b]{0.35\textwidth}
        \centering
        \includegraphics[width=\textwidth]{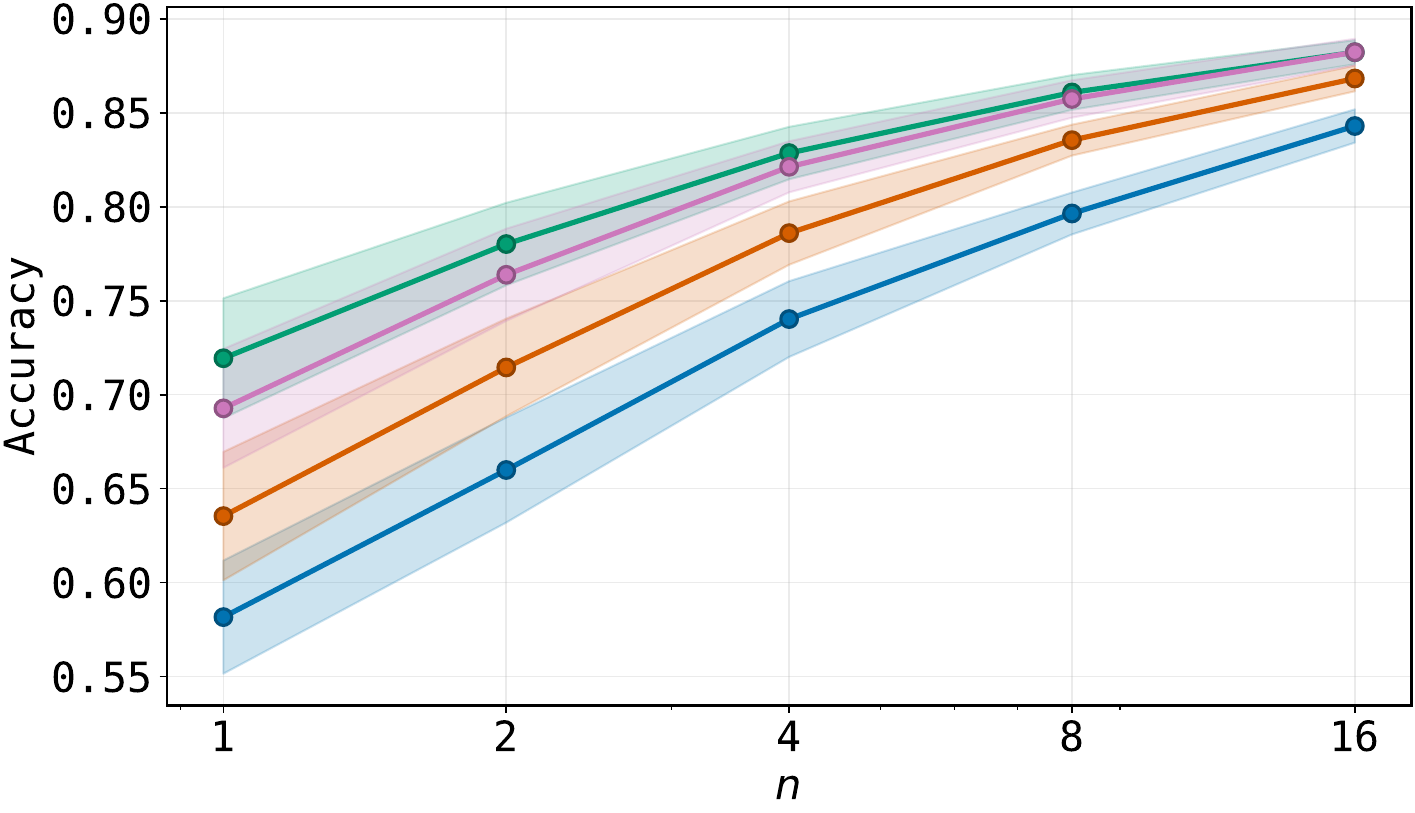}
        \caption{UTO $\rightarrow$ LAB}
    \end{subfigure}
    \hfill
    \begin{subfigure}[b]{0.35\textwidth}
        \centering
        \includegraphics[width=\textwidth]{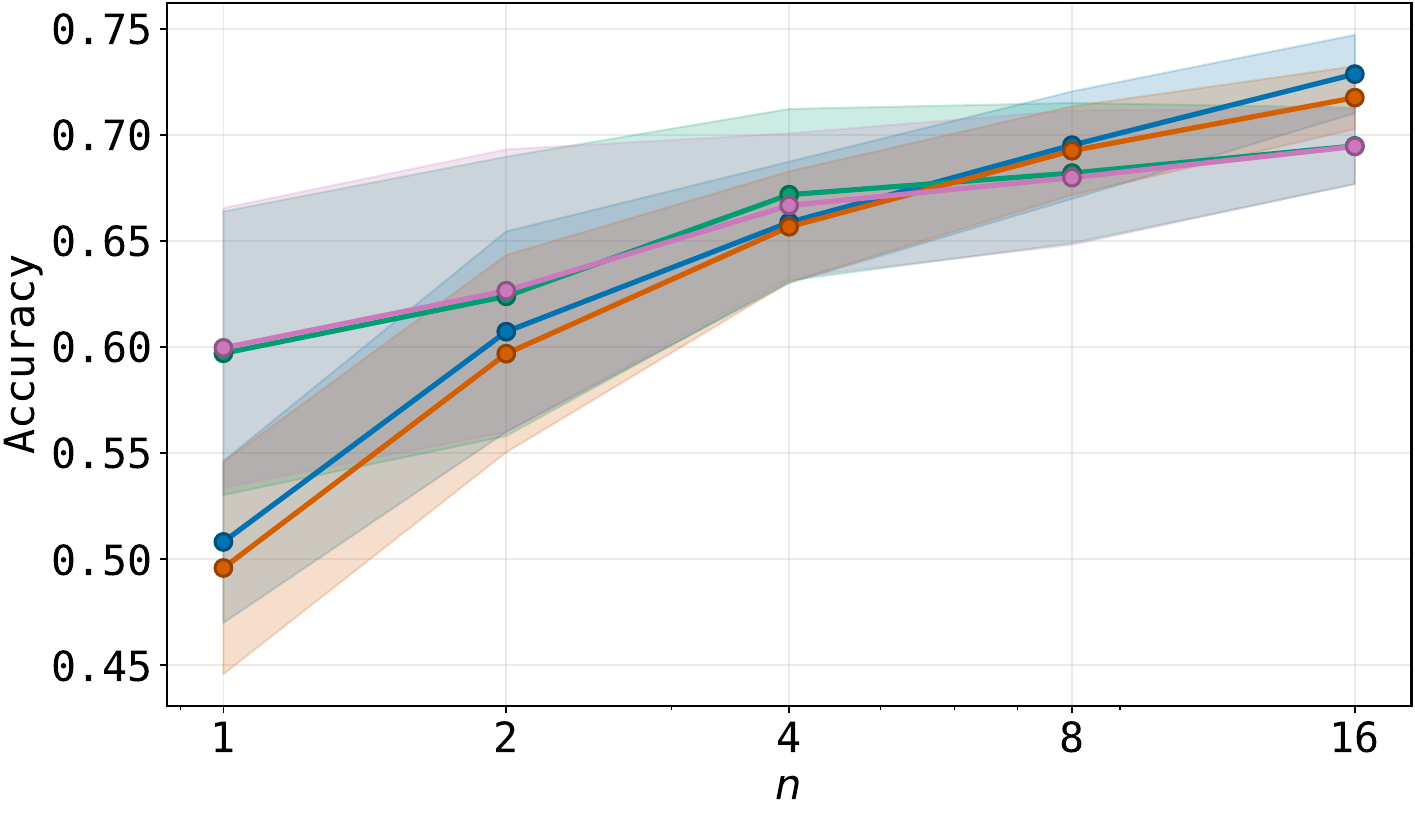}
        \caption{UTO $\rightarrow$ SEA}
    \end{subfigure}
    \hspace{1cm}
     \vskip\baselineskip
    \includegraphics[width=0.7\textwidth]{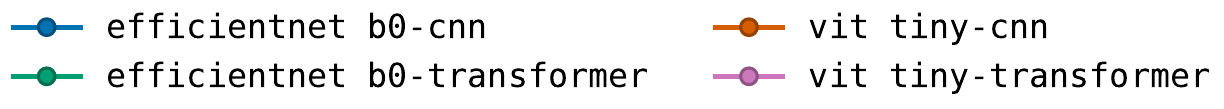}
    \caption{Average accuracy and standard deviation as a function of gallery set size, with $n$ being the number of multimodal samples in the gallery set (I+P $\rightarrow$ I+P)}
    \label{fig:tbd}
\end{figure}

To analyze the impact of gallery set size on recognition accuracy, we evaluated the two best-performing image encoders (EfficientNet-B0 and ViT-Tiny) along with CNN- and Transformer-based profile encoders. Fig.~\ref{fig:tbd} illustrates the mean accuracy and standard deviation as a function of gallery set size, with $n$ being the number of multimodal samples per class. As expected, accuracy improves steadily as the gallery size increases, confirming that larger reference sets improve embedding robustness. Additionally, models with CNN profile encoders perform worse with small gallery sets, but improve faster than transformers. The choice of image encoder architecture has only a minor effect on performance at this scale.

\subsubsection{Comparison with Self-supervised Learning Baseline.}

We compared the performance of our multimodal model against an image-only self-supervised baseline DINO~\cite{caron2021emerging}. DINO was selected as it has been shown to produce strong visual representations directly through self-supervised training without requiring additional fine-tuning~\cite{caron2021emerging}, and due to there not being  suitable multimodal baselines (e.g., BioClip requires images and taxonomic labels, and taxonomic labels are not well defined for plankton).  Detailed training configurations for all models are provided in Appendix~\ref{appendix:A5}. 

Table~\ref{tab:dino_comparison} presents results across all domains and architectures. When evaluated using only image data (I$\rightarrow$I), the multimodal contrastive model consistently outperforms DINO, indicating that multimodal pre-training enhances the quality of visual representations even in unimodal settings. When both image and profile modalities are available during inference (I+P$\rightarrow$I+P), the multimodal model achieves further gains, demonstrating the effectiveness of joint multimodal embeddings for improving domain generalization and overall classification performance.

\begin{table}[tb]
    \centering
    \caption{Average accuracy and standard deviation (\%) comparing DINO and contrastive multimodal pre-training}
    \resizebox{\textwidth}{!}{
    \begin{tabular}{p{2.5cm} p{3cm} *{4}{C{2.5cm}}}
        \toprule
        Method & Model & LAB $\rightarrow$ LAB & LAB $\rightarrow$ SEA & UTO $\rightarrow$ LAB & UTO $\rightarrow$ SEA \\
        \midrule
        \multirow{3}{2cm}{DINO}
        & EfficientNet B0 & 79.67\std{1.42} & 57.30\std{2.29} & 57.24\std{0.81} & 63.19\std{2.29} \\
        & ConvNeXt Femto  & 62.85\std{2.82} & 55.52\std{3.29} & 54.76\std{0.87} & 55.58\std{1.75} \\
        & ViT Tiny        & 59.54\std{2.04} & 54.59\std{1.83} & 53.24\std{1.01} & 59.10\std{2.42} \\
        %\bottomrule
    \end{tabular}}
    \resizebox{\textwidth}{!}{
    \begin{tabular}{p{2.5cm} p{3cm} *{4}{C{2.5cm}}}
        \midrule
        \multirow{3}{2cm}{InfoNCE (I$\rightarrow$I)}
        & EfficientNet B0 & 91.62\std{1.05} & 66.04\std{1.60} & 69.87\std{1.21} & 72.56\std{1.98} \\
        & ConvNeXt Femto  & 90.01\std{0.98} & 65.62\std{1.88} & 66.03\std{1.13} & 72.05\std{1.84} \\
        & ViT Tiny        & 92.79\std{0.55}  & 64.37\std{1.94} & 66.11\std{1.22} & 70.83\std{1.68} \\
        %\bottomrule
    \end{tabular}}
    \resizebox{\textwidth}{!}{
    \begin{tabular}{p{2.5cm} p{3cm} *{4}{C{2.5cm}}}
        \midrule
        \multirow{3}{2cm}{InfoNCE (I+P$\rightarrow$I+P)}
        & EfficientNet B0 & 95.06{\scriptsize\std{0.58}} & 67.57\std{2.23} & \narrowbold{88.35\std{0.70}} & \narrowbold{72.86\std{1.85}} \\
        & ConvNeXt Femto  & 94.81\std{0.90} & \narrowbold{68.05\std{2.06}} & 88.13\std{0.74} & 70.63\std{1.77} \\
        & ViT Tiny        & \narrowbold{96.01\std{0.56}}  & 66.75\std{2.21} & 88.30\std{0.71} & 71.75\std{1.48} \\
        \bottomrule
    \end{tabular}}
    \label{tab:dino_comparison}
\end{table}

\section{Conclusion}
In this work, we explored the benefits of label-free cross-modal contrastive pre-training for plankton recognition. We introduced multimodal datasets and evaluated multiple image and profile encoder combinations and contrastive learning losses under in-domain and cross-domain settings. The results show that multimodal representation coordination significantly improves recognition accuracy compared to single-modality approaches, particularly when both modalities are available during inference.

Beyond improving recognition accuracy, we further demonstrated that pre-training on raw, unlabeled data from diverse environments enables cross-domain generalization, outperforming models trained on controlled laboratory data. These finding highlights the potential of leveraging raw \textit{in-situ} data collected in real-world conditions to reduce reliance on manual annotation and improve scalability. To promote research in multimodal plankton recognition we have made our datasets publicly available~\cite{cytosensedata}.

\begin{credits}
\subsubsection{\ackname} The research was carried out in the FASTVISION and FASTVISION-plus projects funded by the Academy of Finland (Decision numbers 321980, 321991, 339612, and 339355). 

%\subsubsection{\discintname}
%The authors have no competing interests to declare that are relevant to the content of this article.
\end{credits}

\bibliographystyle{splncs04}
\bibliography{ref}

\newpage

\section*{Appendix}

\subsection*{A.1. Profile encoder architectures}
\label{appendix:A1}
The design choices for each encoder were made to balance representational capacity and computational efficiency, given the relatively small dataset size. Detailed configuration parameters for each model are provided below.

\noindent
\textbf{CNN.}
The CNN-based encoder follows the ResNet18 design with the key modification of using only 1D convolutions. The main parameters are summarized in Table~\ref{tab:cnn_params}:

\begin{table}[htb]
    \centering
    \caption{Configuration of the 1D CNN encoder.}
    \begin{tabular}{lc}
        \toprule
        Parameter & Value \\
        \midrule
        Input size & 6 \\
        Output size & 257 \\
        Blocks & [2, 2, 2, 2] \\
        Base channels & 32 \\
        Dropout & 0.2 \\
        \midrule
        Params &  963K\\
        \bottomrule
    \end{tabular}
    \label{tab:cnn_params}
\end{table}

\noindent
\textbf{Bidirectional LSTM.}
The Bidirectional LSTM captures sequential dependencies in both the forward and backward directions. During our experimentation we found that using a single-layer LSTM was sufficient, as adding more layers did not improve performance. The encoder configuration is listed in Table~\ref{tab:lstm_params}.

\begin{table}[htb]
    \centering
    \caption{Configuration of the Bidirectional LSTM encoder.}
    \begin{tabular}{lc}
        \toprule
        Parameter & Value \\
        \midrule
        Input size & 6 \\
        Output size & 257 \\
        Hidden size & 128 \\
        N layers & 1 \\
        Dropout & 0.2 \\
        \midrule
        Params & 264K \\
        \bottomrule
    \end{tabular}
    \label{tab:lstm_params}
\end{table}

\noindent
\textbf{Transformer.}
The Transformer encoder adopted a lightweight configuration inspired by ViT-Tiny, consisting of six encoder layers with four attention heads per layer and GELU activations in the feedforward layers. Table~\ref{tab:transformer_params} summarizes the parameters.

\begin{table}[htb]
    \centering
    \caption{Configuration of the Transformer encoder.}
    \begin{tabular}{lc}
        \toprule
        Parameter & Value \\
        \midrule
        Input size & 6 \\
        Output size & 129 \\
        Hidden size & 128 \\
        N heads & 4 \\
        N layers & 6 \\
        Feedforward Dim & 256 \\
        Dropout & 0.2 \\
        \midrule
        Params & 1.1M\\
        \bottomrule
    \end{tabular}
    \label{tab:transformer_params}
\end{table}

\subsection*{A.2. In-domain Evaluation Full Results}
\label{appendix:A2}
The full results for the in-domain evaluation with all encoder pairs with InfoNCE loss can be seen in Table \ref{tab:lab_lab_infonce_full} and for Sigmoid loss in Table \ref{tab:lab_lab_sig_full}. Notably, InfoNCE does better in all cases. When comparing \gls{lstm} to the other two encoders, the results show that especially in I$\rightarrow$P and P$\rightarrow$I, \gls{lstm} attains much lower accuracy.

\begin{table}[htb]
    \centering
    \caption{Full results, LAB $\rightarrow$ LAB, InfoNCE}
    \resizebox{.32\textwidth}{!}{%
    \begin{tabular}{C{1.5cm} C{1.5cm} C{2cm}}
        \toprule
        \multicolumn{3}{c}{EfficientNet B0 + CNN} \\
        \midrule
        Gallery & Test & Accuracy \\
        \midrule
        I    & I    & 89.76$\pm$0.65\% \\
        I    & P    & 90.95$\pm$0.80\% \\
        I    & I+P  & 92.01$\pm$0.68\% \\
        P    & I    & 89.05$\pm$0.83\% \\
        P    & P    & 93.93$\pm$0.69\% \\
        P    & I+P  & 94.65$\pm$0.68\% \\
        I+P  & I    & 89.82$\pm$0.64\% \\
        I+P  & P    & 93.97$\pm$0.71\% \\
        I+P  & I+P  & 95.06$\pm$0.58\% \\
        \bottomrule
    \end{tabular}%
    }
    \hfill
    \resizebox{.32\textwidth}{!}{%
    \begin{tabular}{C{1.5cm} C{1.5cm} C{2cm}}
        \toprule
        \multicolumn{3}{c}{ConvNeXt Femto + CNN} \\
        \midrule
        Gallery & Test & Accuracy \\
        \midrule
        I    & I    & 89.21$\pm$2.57\% \\
        I    & P    & 91.27$\pm$1.19\% \\
        I    & I+P  & 91.68$\pm$1.96\% \\
        P    & I    & 89.09$\pm$1.99\% \\
        P    & P    & 93.73$\pm$0.68\% \\
        P    & I+P  & 94.41$\pm$0.78\% \\
        I+P  & I    & 89.26$\pm$2.58\% \\
        I+P  & P    & 93.76$\pm$0.66\% \\
        I+P  & I+P  & 94.81$\pm$0.90\% \\
        \bottomrule
    \end{tabular}%
    }
    \hfill
    \resizebox{.32\textwidth}{!}{%
    \begin{tabular}{C{1.5cm} C{1.5cm} C{2cm}}
        \toprule
        \multicolumn{3}{c}{ViT-Tiny + CNN} \\
        \midrule
        Gallery & Test & Accuracy \\
        \midrule
        I    & I    & 92.73$\pm$0.59\% \\
        I    & P    & 93.03$\pm$0.69\% \\
        I    & I+P  & 94.24$\pm$0.59\% \\
        P    & I    & 91.56$\pm$0.67\% \\
        P    & P    & 94.51$\pm$0.83\% \\
        P    & I+P  & 95.42$\pm$0.73\% \\
        I+P  & I    & 92.79$\pm$0.57\% \\
        I+P  & P    & 94.55$\pm$0.82\% \\
        I+P  & I+P  & 96.01$\pm$0.56\% \\
        \bottomrule
    \end{tabular}%
    }
    \hfill
    \resizebox{.32\textwidth}{!}{%
    \begin{tabular}{C{1.5cm} C{1.5cm} C{2cm}}
        \toprule
        \multicolumn{3}{c}{EfficientNet B0 + LSTM} \\
        \midrule
        Gallery & Test & Accuracy \\
        \midrule
        I    & I    & 89.78$\pm$0.71\% \\
        I    & P    & 76.62$\pm$1.43\% \\
        I    & I+P  & 90.76$\pm$0.73\% \\
        P    & I    & 80.23$\pm$1.11\% \\
        P    & P    & 82.60$\pm$0.90\% \\
        P    & I+P  & 83.95$\pm$0.81\% \\
        I+P  & I    & 89.75$\pm$0.71\% \\
        I+P  & P    & 82.60$\pm$0.90\% \\
        I+P  & I+P  & 89.59$\pm$0.67\% \\
        \bottomrule
    \end{tabular}%
    }
    \hfill
    \resizebox{.32\textwidth}{!}{%
    \begin{tabular}{C{1.5cm} C{1.5cm} C{2cm}}
        \toprule
        \multicolumn{3}{c}{ConvNeXt Femto + LSTM} \\
        \midrule
        Gallery & Test & Accuracy \\
        \midrule
        I    & I    & 90.01$\pm$0.98\% \\
        I    & P    & 78.46$\pm$1.00\% \\
        I    & I+P  & 90.92$\pm$0.85\% \\
        P    & I    & 81.20$\pm$1.17\% \\
        P    & P    & 81.84$\pm$0.86\% \\
        P    & I+P  & 83.59$\pm$0.80\% \\
        I+P  & I    & 90.01$\pm$0.99\% \\
        I+P  & P    & 81.84$\pm$0.85\% \\
        I+P  & I+P  & 90.10$\pm$0.67\% \\
        \bottomrule
    \end{tabular}%
    }
    \hfill
    \resizebox{.32\textwidth}{!}{%
    \begin{tabular}{C{1.5cm} C{1.5cm} C{2cm}}
        \toprule
        \multicolumn{3}{c}{ViT-Tiny + LSTM} \\
        \midrule
        Gallery & Test & Accuracy \\
        \midrule
        I    & I    & 90.95$\pm$0.67\% \\
        I    & P    & 78.76$\pm$1.44\% \\
        I    & I+P  & 91.64$\pm$0.63\% \\
        P    & I    & 82.11$\pm$0.93\% \\
        P    & P    & 81.64$\pm$0.87\% \\
        P    & I+P  & 83.39$\pm$0.81\% \\
        I+P  & I    & 90.95$\pm$0.67\% \\
        I+P  & P    & 81.65$\pm$0.87\% \\
        I+P  & I+P  & 90.39$\pm$0.66\% \\
        \bottomrule
    \end{tabular}%
    }
    \hfill
    \resizebox{.32\textwidth}{!}{%
    \begin{tabular}{C{1.5cm} C{1.5cm} C{2cm}}
        \toprule
        \multicolumn{3}{c}{EfficientNet B0 + Transformer} \\
        \midrule
        Gallery & Test & Accuracy \\
        \midrule
        I    & I    & 91.62$\pm$1.05\% \\
        I    & P    & 86.64$\pm$1.70\% \\
        I    & I+P  & 92.61$\pm$0.98\% \\
        P    & I    & 87.41$\pm$1.03\% \\
        P    & P    & 88.94$\pm$0.97\% \\
        P    & I+P  & 89.60$\pm$1.04\% \\
        I+P  & I    & 91.64$\pm$1.05\% \\
        I+P  & P    & 88.94$\pm$0.97\% \\
        I+P  & I+P  & 92.11$\pm$0.98\% \\
        \bottomrule
    \end{tabular}%
    }
    \hfill
    \resizebox{.32\textwidth}{!}{%
    \begin{tabular}{C{1.5cm} C{1.5cm} C{2cm}}
        \toprule
        \multicolumn{3}{c}{ConvNeXt Femto + Transformer} \\
        \midrule
        Gallery & Test & Accuracy \\
        \midrule
        I    & I    & 88.67$\pm$6.48\% \\
        I    & P    & 79.97$\pm$8.21\% \\
        I    & I+P  & 89.72$\pm$6.21\% \\
        P    & I    & 82.13$\pm$10.0\% \\
        P    & P    & 86.28$\pm$3.60\% \\
        P    & I+P  & 86.84$\pm$3.75\% \\
        I+P  & I    & 88.67$\pm$6.48\% \\
        I+P  & P    & 86.28$\pm$3.60\% \\
        I+P  & I+P  & 89.85$\pm$3.84\% \\
        \bottomrule
    \end{tabular}%
    } 
    \hfill
    \resizebox{.32\textwidth}{!}{%
    \begin{tabular}{C{1.5cm} C{1.5cm} C{2cm}}
        \toprule
        \multicolumn{3}{c}{ViT-Tiny + Transformer} \\
        \midrule
        Gallery & Test & Accuracy \\
        \midrule
        I    & I    & 92.79$\pm$0.55\% \\
        I    & P    & 85.36$\pm$1.24\% \\
        I    & I+P  & 93.33$\pm$0.54\% \\
        P    & I    & 87.58$\pm$1.18\% \\
        P    & P    & 87.71$\pm$0.84\% \\
        P    & I+P  & 88.37$\pm$0.82\% \\
        I+P  & I    & 92.79$\pm$0.55\% \\
        I+P  & P    & 87.71$\pm$0.84\% \\
        I+P  & I+P  & 91.80$\pm$0.76\% \\
        \bottomrule
    \end{tabular}%
    } 
    \label{tab:lab_lab_infonce_full}
\end{table}

% % % % % % % % % % % % % % % % % % % % % % % % 
% LAB -> LAB, Sigmoid
% % % % % % % % % % % % % % % % % % % % % % % % 
\begin{table}[htb]
    \centering
    \caption{Full results, LAB $\rightarrow$ LAB, Sigmoid}
    \resizebox{.32\textwidth}{!}{%
    \begin{tabular}{C{1.5cm} C{1.5cm} C{2cm}}
        \toprule
        \multicolumn{3}{c}{EfficientNet B0 + CNN} \\
        \midrule
        Gallery & Test & Accuracy \\
        \midrule
        I    & I    & 80.58$\pm$1.32\% \\
        I    & P    & 84.41$\pm$1.37\% \\
        I    & I+P  & 83.52$\pm$1.29\% \\
        P    & I    & 79.55$\pm$1.20\% \\
        P    & P    & 90.79$\pm$0.63\% \\
        P    & I+P  & 91.10$\pm$0.64\% \\
        I+P  & I    & 80.58$\pm$1.32\% \\
        I+P  & P    & 90.79$\pm$0.63\% \\
        I+P  & I+P  & 91.85$\pm$0.65\% \\
        \bottomrule
    \end{tabular}%
    }
    \hfill
    \resizebox{.32\textwidth}{!}{%
    \begin{tabular}{C{1.5cm} C{1.5cm} C{2cm}}
        \toprule
        \multicolumn{3}{c}{ConvNeXt Femto + CNN} \\
        \midrule
        Gallery & Test & Accuracy \\
        \midrule
        I    & I    & 75.25$\pm$3.76\% \\
        I    & P    & 80.74$\pm$3.66\% \\
        I    & I+P  & 77.43$\pm$3.83\% \\
        P    & I    & 73.26$\pm$4.25\% \\
        P    & P    & 89.46$\pm$1.68\% \\
        P    & I+P  & 89.67$\pm$1.65\% \\
        I+P  & I    & 75.25$\pm$3.76\% \\
        I+P  & P    & 89.46$\pm$1.68\% \\
        I+P  & I+P  & 89.93$\pm$1.81\% \\
        \bottomrule
    \end{tabular}%
    }
    \hfill
    \resizebox{.32\textwidth}{!}{%
    \begin{tabular}{C{1.5cm} C{1.5cm} C{2cm}}
        \toprule
        \multicolumn{3}{c}{ViT-Tiny + CNN} \\
        \midrule
        Gallery & Test & Accuracy \\
        \midrule
        I    & I    & 90.51$\pm$0.73\% \\
        I    & P    & 91.32$\pm$0.87\% \\
        I    & I+P  & 91.67$\pm$0.64\% \\
        P    & I    & 88.97$\pm$0.77\% \\
        P    & P    & 93.35$\pm$0.54\% \\
        P    & I+P  & 93.68$\pm$0.56\% \\
        I+P  & I    & 90.51$\pm$0.73\% \\
        I+P  & P    & 93.35$\pm$0.54\% \\
        I+P  & I+P  & 94.89$\pm$0.43\% \\
        \bottomrule
    \end{tabular}%
    }
    \hfill
    \resizebox{.32\textwidth}{!}{%
    \begin{tabular}{C{1.5cm} C{1.5cm} C{2cm}}
        \toprule
        \multicolumn{3}{c}{EfficientNet B0 + LSTM} \\
        \midrule
        Gallery & Test & Accuracy \\
        \midrule
        I    & I    & 82.11$\pm$0.93\% \\
        I    & P    & 60.38$\pm$1.43\% \\
        I    & I+P  & 83.75$\pm$0.80\% \\
        P    & I    & 63.22$\pm$1.73\% \\
        P    & P    & 75.23$\pm$1.11\% \\
        P    & I+P  & 76.74$\pm$1.02\% \\
        I+P  & I    & 82.10$\pm$0.93\% \\
        I+P  & P    & 75.22$\pm$1.11\% \\
        I+P  & I+P  & 84.13$\pm$0.77\% \\
        \bottomrule
    \end{tabular}%
    }
    \hfill
    \resizebox{.32\textwidth}{!}{%
    \begin{tabular}{C{1.5cm} C{1.5cm} C{2cm}}
        \toprule
        \multicolumn{3}{c}{ConvNeXt Femto + LSTM} \\
        \midrule
        Gallery & Test & Accuracy \\
        \midrule
        I    & I    & 84.14$\pm$3.93\% \\
        I    & P    & 63.33$\pm$2.56\% \\
        I    & I+P  & 85.43$\pm$3.47\% \\
        P    & I    & 65.39$\pm$5.04\% \\
        P    & P    & 74.81$\pm$1.03\% \\
        P    & I+P  & 76.66$\pm$1.02\% \\
        I+P  & I    & 84.14$\pm$3.93\% \\
        I+P  & P    & 74.81$\pm$1.03\% \\
        I+P  & I+P  & 85.82$\pm$1.50\% \\
        \bottomrule
    \end{tabular}%
    }
    \hfill
    \resizebox{.32\textwidth}{!}{%
    \begin{tabular}{C{1.5cm} C{1.5cm} C{2cm}}
        \toprule
        \multicolumn{3}{c}{ViT-Tiny + LSTM} \\
        \midrule
        Gallery & Test & Accuracy \\
        \midrule
        I    & I    & 87.93$\pm$1.53\% \\
        I    & P    & 64.56$\pm$1.78\% \\
        I    & I+P  & 88.69$\pm$1.36\% \\
        P    & I    & 69.35$\pm$2.17\% \\
        P    & P    & 74.63$\pm$1.19\% \\
        P    & I+P  & 76.45$\pm$1.13\% \\
        I+P  & I    & 87.93$\pm$1.53\% \\
        I+P  & P    & 74.63$\pm$1.19\% \\
        I+P  & I+P  & 87.45$\pm$1.15\% \\
        \bottomrule
    \end{tabular}%
    }
    \hfill
    \resizebox{.32\textwidth}{!}{%
    \begin{tabular}{C{1.5cm} C{1.5cm} C{2cm}}
        \toprule
        \multicolumn{3}{c}{EfficientNet B0 + Transformer} \\
        \midrule
        Gallery & Test & Accuracy \\
        \midrule
        I    & I    & 81.54$\pm$0.93\% \\
        I    & P    & 73.33$\pm$1.81\% \\
        I    & I+P  & 83.47$\pm$0.94\% \\
        P    & I    & 71.46$\pm$1.90\% \\
        P    & P    & 84.04$\pm$1.44\% \\
        P    & I+P  & 84.34$\pm$1.43\% \\
        I+P  & I    & 81.55$\pm$0.93\% \\
        I+P  & P    & 84.04$\pm$1.44\% \\
        I+P  & I+P  & 85.98$\pm$1.28\% \\
        \bottomrule
    \end{tabular}%
    }
    \hfill
    \resizebox{.32\textwidth}{!}{%
    \begin{tabular}{C{1.5cm} C{1.5cm} C{2cm}}
        \toprule
        \multicolumn{3}{c}{ConvNeXt Femto + Transformer} \\
        \midrule
        Gallery & Test & Accuracy \\
        \midrule
        I    & I    & 83.48$\pm$8.13\% \\
        I    & P    & 75.13$\pm$6.19\% \\
        I    & I+P  & 84.68$\pm$7.51\% \\
        P    & I    & 73.02$\pm$9.00\% \\
        P    & P    & 83.38$\pm$2.32\% \\
        P    & I+P  & 83.70$\pm$2.36\% \\
        I+P  & I    & 83.48$\pm$8.13\% \\
        I+P  & P    & 83.38$\pm$2.32\% \\
        I+P  & I+P  & 86.56$\pm$2.54\% \\
        \bottomrule
    \end{tabular}%
    } 
    \hfill
    \resizebox{.32\textwidth}{!}{%
    \begin{tabular}{C{1.5cm} C{1.5cm} C{2cm}}
        \toprule
        \multicolumn{3}{c}{ViT-Tiny + Transformer} \\
        \midrule
        Gallery & Test & Accuracy \\
        \midrule
        I    & I    & 90.42$\pm$0.80\% \\
        I    & P    & 81.21$\pm$1.98\% \\
        I    & I+P  & 91.04$\pm$0.83\% \\
        P    & I    & 82.19$\pm$2.89\% \\
        P    & P    & 86.41$\pm$1.91\% \\
        P    & I+P  & 86.88$\pm$1.87\% \\
        I+P  & I    & 90.42$\pm$0.80\% \\
        I+P  & P    & 86.41$\pm$1.91\% \\
        I+P  & I+P  & 90.24$\pm$1.27\% \\
        \bottomrule
    \end{tabular}%
    } 
    \label{tab:lab_lab_sig_full}
\end{table}

\subsection*{A.3. Cross-domain Evaluation Full Results}
\label{appendix:A3}
Full results for the cross-domain evaluation, using all image encoders and the CNN/Transformer profile encoder with InfoNCE loss are presented in Tables \ref{tab:lab_sea_full}, \ref{tab:uto_lab_full}, \ref{tab:uto_sea_full} for LAB$\rightarrow$SEA, UTO$\rightarrow$LAB, and UTO$\rightarrow$SEA respectively. Notably, in the cross-domain setting, the cross-modal performance (I$\rightarrow$P, P$\rightarrow$I) is significantly worse than the intramodal performance (I$\rightarrow$I, P$\rightarrow$P). This result seems to indicate that with more realistic profile data the model does not actually manage to coordinate the representations, possibly due to the low amount of training data used or the long-tailed nature of the data. However, despite the low cross-modal performance, the model achieves very high accuracy when the modalities are used either independently or combined.

% % % % % % % % % % % % % % % % % % % % % % % % 
% LAB -> SEA, InfoNCE
% % % % % % % % % % % % % % % % % % % % % % % % 
\begin{table}[htb]
    \centering
    \caption{Full results, LAB $\rightarrow$ SEA, InfoNCE}
    \resizebox{.32\textwidth}{!}{%
    \begin{tabular}{C{1.5cm} C{1.5cm} C{2cm}}
        \toprule
        \multicolumn{3}{c}{EfficientNet B0 + CNN} \\
        \midrule
        Gallery & Test & Accuracy \\
        \midrule
        I    & I    & 63.47$\pm$2.47\% \\
        I    & P    & 29.35$\pm$3.80\% \\
        I    & I+P  & 64.25$\pm$2.72\% \\
        P    & I    & 28.58$\pm$4.75\% \\
        P    & P    & 60.31$\pm$2.21\% \\
        P    & I+P  & 61.11$\pm$2.34\% \\
        I+P  & I    & 63.39$\pm$2.48\% \\
        I+P  & P    & 60.33$\pm$2.20\% \\
        I+P  & I+P  & 67.57$\pm$2.23\% \\
        \bottomrule
    \end{tabular}%
    }
    \hfill
    \resizebox{.32\textwidth}{!}{%
    \begin{tabular}{C{1.5cm} C{1.5cm} C{2cm}}
        \toprule
        \multicolumn{3}{c}{ConvNeXt Femto + CNN} \\
        \midrule
        Gallery & Test & Accuracy \\
        \midrule
        I    & I    & 62.84$\pm$2.12\% \\
        I    & P    & 30.64$\pm$3.96\% \\
        I    & I+P  & 63.80$\pm$2.70\% \\
        P    & I    & 28.99$\pm$3.87\% \\
        P    & P    & 60.71$\pm$2.38\% \\
        P    & I+P  & 61.49$\pm$2.46\% \\
        I+P  & I    & 62.76$\pm$2.12\% \\
        I+P  & P    & 60.72$\pm$2.37\% \\
        I+P  & I+P  & 68.05$\pm$2.06\% \\
        \bottomrule
    \end{tabular}%
    }
    \hfill
    \resizebox{.32\textwidth}{!}{%
    \begin{tabular}{C{1.5cm} C{1.5cm} C{2cm}}
        \toprule
        \multicolumn{3}{c}{ViT-Tiny + CNN} \\
        \midrule
        Gallery & Test & Accuracy \\
        \midrule
        I    & I    & 60.59$\pm$2.15\% \\
        I    & P    & 29.64$\pm$4.08\% \\
        I    & I+P  & 61.63$\pm$2.68\% \\
        P    & I    & 25.94$\pm$4.23\% \\
        P    & P    & 59.59$\pm$2.07\% \\
        P    & I+P  & 60.19$\pm$2.09\% \\
        I+P  & I    & 60.57$\pm$2.15\% \\
        I+P  & P    & 59.61$\pm$2.06\% \\
        I+P  & I+P  & 66.75$\pm$2.21\% \\
        \bottomrule
    \end{tabular}%
    }
    \hfill
    \resizebox{.32\textwidth}{!}{%
    \begin{tabular}{C{1.5cm} C{1.5cm} C{2cm}}
        \toprule
        \multicolumn{3}{c}{EfficientNet B0 + Transformer} \\
        \midrule
        Gallery & Test & Accuracy \\
        \midrule
        I    & I    & 66.04$\pm$1.60\% \\
        I    & P    & 24.80$\pm$4.79\% \\
        I    & I+P  & 66.28$\pm$1.77\% \\
        P    & I    & 17.66$\pm$4.12\% \\
        P    & P    & 61.88$\pm$2.16\% \\
        P    & I+P  & 62.01$\pm$2.19\% \\
        I+P  & I    & 66.04$\pm$1.60\% \\
        I+P  & P    & 61.88$\pm$2.16\% \\
        I+P  & I+P  & 66.87$\pm$2.07\% \\
        \bottomrule
    \end{tabular}%
    }
    \hfill
    \resizebox{.32\textwidth}{!}{%
    \begin{tabular}{C{1.5cm} C{1.5cm} C{2cm}}
        \toprule
        \multicolumn{3}{c}{ConvNeXt Femto + Transformer} \\
        \midrule
        Gallery & Test & Accuracy \\
        \midrule
        I    & I    & 65.62$\pm$1.88\% \\
        I    & P    & 21.98$\pm$5.42\% \\
        I    & I+P  & 65.71$\pm$1.87\% \\
        P    & I    & 16.36$\pm$4.89\% \\
        P    & P    & 59.71$\pm$3.26\% \\
        P    & I+P  & 59.81$\pm$3.30\% \\
        I+P  & I    & 65.60$\pm$1.89\% \\
        I+P  & P    & 59.71$\pm$3.26\% \\
        I+P  & I+P  & 64.49$\pm$3.42\% \\
        \bottomrule
    \end{tabular}%
    } 
    \hfill
    \resizebox{.32\textwidth}{!}{%
    \begin{tabular}{C{1.5cm} C{1.5cm} C{2cm}}
        \toprule
        \multicolumn{3}{c}{ViT-Tiny + Transformer} \\
        \midrule
        Gallery & Test & Accuracy \\
        \midrule
        I    & I    & 64.37$\pm$1.94\% \\
        I    & P    & 25.41$\pm$5.45\% \\
        I    & I+P  & 64.80$\pm$2.04\% \\
        P    & I    & 17.98$\pm$3.03\% \\
        P    & P    & 61.41$\pm$2.00\% \\
        P    & I+P  & 61.53$\pm$2.00\% \\
        I+P  & I    & 64.36$\pm$1.94\% \\
        I+P  & P    & 61.41$\pm$2.00\% \\
        I+P  & I+P  & 66.42$\pm$1.90\% \\
        \bottomrule
    \end{tabular}%
    }
    \label{tab:lab_sea_full}
\end{table}

% % % % % % % % % % % % % % % % % % % % % % % % 
% UTO -> LAB, InfoNCE
% % % % % % % % % % % % % % % % % % % % % % % % 
\begin{table}[htb]
    \centering
    \caption{Full results, UTO $\rightarrow$ LAB, InfoNCE}
    \resizebox{.32\textwidth}{!}{%
    \begin{tabular}{C{1.5cm} C{1.5cm} C{2cm}}
        \toprule
        \multicolumn{3}{c}{EfficientNet B0 + CNN} \\
        \midrule
        Gallery & Test & Accuracy \\
        \midrule
        I    & I    & 69.87$\pm$1.21\% \\
        I    & P    & 17.54$\pm$2.17\% \\
        I    & I+P  & 69.81$\pm$1.21\% \\
        P    & I    & 16.35$\pm$1.20\% \\
        P    & P    & 80.01$\pm$0.95\% \\
        P    & I+P  & 79.96$\pm$1.00\% \\
        I+P  & I    & 69.85$\pm$1.21\% \\
        I+P  & P    & 80.01$\pm$0.95\% \\
        I+P  & I+P  & 84.22$\pm$0.99\% \\
        \bottomrule
    \end{tabular}%
    }
    \hfill
    \resizebox{.32\textwidth}{!}{%
    \begin{tabular}{C{1.5cm} C{1.5cm} C{2cm}}
        \toprule
        \multicolumn{3}{c}{ConvNeXt Femto + CNN} \\
        \midrule
        Gallery & Test & Accuracy \\
        \midrule
        I    & I    & 66.03$\pm$1.13\% \\
        I    & P    & 16.26$\pm$2.11\% \\
        I    & I+P  & 66.15$\pm$1.22\% \\
        P    & I    & 17.09$\pm$1.06\% \\
        P    & P    & 80.97$\pm$0.93\% \\
        P    & I+P  & 80.98$\pm$0.96\% \\
        I+P  & I    & 66.01$\pm$1.14\% \\
        I+P  & P    & 80.96$\pm$0.92\% \\
        I+P  & I+P  & 83.73$\pm$0.79\% \\
        \bottomrule
    \end{tabular}%
    }
    \hfill
    \resizebox{.32\textwidth}{!}{%
    \begin{tabular}{C{1.5cm} C{1.5cm} C{2cm}}
        \toprule
        \multicolumn{3}{c}{ViT-Tiny + CNN} \\
        \midrule
        Gallery & Test & Accuracy \\
        \midrule
        I    & I    & 66.11$\pm$1.22\% \\
        I    & P    & 13.97$\pm$2.31\% \\
        I    & I+P  & 65.93$\pm$1.22\% \\
        P    & I    & 16.27$\pm$1.32\% \\
        P    & P    & 84.30$\pm$0.90\% \\
        P    & I+P  & 84.38$\pm$0.91\% \\
        I+P  & I    & 66.10$\pm$1.22\% \\
        I+P  & P    & 84.31$\pm$0.89\% \\
        I+P  & I+P  & 86.81$\pm$0.83\% \\
        \bottomrule
    \end{tabular}%
    }
    \hfill
    \resizebox{.32\textwidth}{!}{%
    \begin{tabular}{C{1.5cm} C{1.5cm} C{2cm}}
        \toprule
        \multicolumn{3}{c}{EfficientNet B0 + Transformer} \\
        \midrule
        Gallery & Test & Accuracy \\
        \midrule
        I    & I    & 69.43$\pm$1.10\% \\
        I    & P    & 15.96$\pm$2.96\% \\
        I    & I+P  & 69.08$\pm$1.07\% \\
        P    & I    & 18.32$\pm$0.78\% \\
        P    & P    & 87.33$\pm$0.79\% \\
        P    & I+P  & 87.36$\pm$0.79\% \\
        I+P  & I    & 69.43$\pm$1.09\% \\
        I+P  & P    & 87.33$\pm$0.79\% \\
        I+P  & I+P  & 88.35$\pm$0.70\% \\
        \bottomrule
    \end{tabular}%
    }
    \hfill
    \resizebox{.32\textwidth}{!}{%
    \begin{tabular}{C{1.5cm} C{1.5cm} C{2cm}}
        \toprule
        \multicolumn{3}{c}{ConvNeXt Femto + Transformer} \\
        \midrule
        Gallery & Test & Accuracy \\
        \midrule
        I    & I    & 66.06$\pm$0.94\% \\
        I    & P    & 15.96$\pm$2.84\% \\
        I    & I+P  & 65.59$\pm$0.97\% \\
        P    & I    & 14.20$\pm$0.79\% \\
        P    & P    & 87.02$\pm$0.72\% \\
        P    & I+P  & 87.04$\pm$0.75\% \\
        I+P  & I    & 66.06$\pm$0.94\% \\
        I+P  & P    & 87.02$\pm$0.72\% \\
        I+P  & I+P  & 88.13$\pm$0.74\% \\
        \bottomrule
    \end{tabular}%
    }
    \hfill
    \resizebox{.32\textwidth}{!}{%
    \begin{tabular}{C{1.5cm} C{1.5cm} C{2cm}}
        \toprule
        \multicolumn{3}{c}{ViT-Tiny + Transformer} \\
        \midrule
        Gallery & Test & Accuracy \\
        \midrule
        I    & I    & 62.41$\pm$1.22\% \\
        I    & P    & 16.88$\pm$2.68\% \\
        I    & I+P  & 62.43$\pm$1.26\% \\
        P    & I    & 15.03$\pm$0.71\% \\
        P    & P    & 87.45$\pm$0.71\% \\
        P    & I+P  & 87.46$\pm$0.71\% \\
        I+P  & I    & 62.41$\pm$1.22\% \\
        I+P  & P    & 87.45$\pm$0.71\% \\
        I+P  & I+P  & 88.30$\pm$0.71\% \\
        \bottomrule
    \end{tabular}%
    } 
    \label{tab:uto_lab_full}
\end{table}

% % % % % % % % % % % % % % % % % % % % % % % % 
% UTO -> SEA, InfoNCE
% % % % % % % % % % % % % % % % % % % % % % % % 
\begin{table}[htb]
    \centering
    \caption{Full results, UTO $\rightarrow$ SEA, InfoNCE}
    \resizebox{.32\textwidth}{!}{%
    \begin{tabular}{C{1.5cm} C{1.5cm} C{2cm}}
        \toprule
        \multicolumn{3}{c}{EfficientNet B0 + CNN} \\
        \midrule
        Gallery & Test & Accuracy \\
        \midrule
        I    & I    & 71.04$\pm$2.01\% \\
        I    & P    & 31.37$\pm$3.94\% \\
        I    & I+P  & 71.69$\pm$2.14\% \\
        P    & I    & 21.49$\pm$2.01\% \\
        P    & P    & 64.73$\pm$2.15\% \\
        P    & I+P  & 64.98$\pm$2.17\% \\
        I+P  & I    & 70.77$\pm$2.02\% \\
        I+P  & P    & 64.73$\pm$2.15\% \\
        I+P  & I+P  & 72.86$\pm$1.85\% \\
        \bottomrule
    \end{tabular}%
    }
    \hfill
    \resizebox{.32\textwidth}{!}{%
    \begin{tabular}{C{1.5cm} C{1.5cm} C{2cm}}
        \toprule
        \multicolumn{3}{c}{ConvNeXt Femto + CNN} \\
        \midrule
        Gallery & Test & Accuracy \\
        \midrule
        I    & I    & 66.38$\pm$2.48\% \\
        I    & P    & 23.32$\pm$3.72\% \\
        I    & I+P  & 65.96$\pm$2.83\% \\
        P    & I    & 22.94$\pm$3.82\% \\
        P    & P    & 63.62$\pm$1.96\% \\
        P    & I+P  & 63.54$\pm$2.03\% \\
        I+P  & I    & 66.07$\pm$2.52\% \\
        I+P  & P    & 63.62$\pm$1.96\% \\
        I+P  & I+P  & 70.63$\pm$1.77\% \\
        \bottomrule
    \end{tabular}%
    }
    \hfill
    \resizebox{.32\textwidth}{!}{%
    \begin{tabular}{C{1.5cm} C{1.5cm} C{2cm}}
        \toprule
        \multicolumn{3}{c}{ViT-Tiny + CNN} \\
        \midrule
        Gallery & Test & Accuracy \\
        \midrule
        I    & I    & 70.83$\pm$1.68\% \\
        I    & P    & 25.61$\pm$3.36\% \\
        I    & I+P  & 70.91$\pm$1.70\% \\
        P    & I    & 25.12$\pm$4.67\% \\
        P    & P    & 62.52$\pm$2.25\% \\
        P    & I+P  & 62.62$\pm$2.35\% \\
        I+P  & I    & 70.79$\pm$1.70\% \\
        I+P  & P    & 62.52$\pm$2.24\% \\
        I+P  & I+P  & 71.75$\pm$1.48\% \\
        \bottomrule
    \end{tabular}%
    }
    \hfill
    \resizebox{.32\textwidth}{!}{%
    \begin{tabular}{C{1.5cm} C{1.5cm} C{2cm}}
        \toprule
        \multicolumn{3}{c}{EfficientNet B0 + Transformer} \\
        \midrule
        Gallery & Test & Accuracy \\
        \midrule
        I    & I    & 72.56$\pm$1.98\% \\
        I    & P    & 38.77$\pm$7.68\% \\
        I    & I+P  & 73.08$\pm$1.70\% \\
        P    & I    & 23.43$\pm$3.21\% \\
        P    & P    & 64.55$\pm$1.93\% \\
        P    & I+P  & 64.64$\pm$1.93\% \\
        I+P  & I    & 72.55$\pm$1.99\% \\
        I+P  & P    & 64.55$\pm$1.93\% \\
        I+P  & I+P  & 69.04$\pm$2.01\% \\
        \bottomrule
    \end{tabular}%
    }
    \hfill
    \resizebox{.32\textwidth}{!}{%
    \begin{tabular}{C{1.5cm} C{1.5cm} C{2cm}}
        \toprule
        \multicolumn{3}{c}{ConvNeXt Femto + Transformer} \\
        \midrule
        Gallery & Test & Accuracy \\
        \midrule
        I    & I    & 72.05$\pm$1.84\% \\
        I    & P    & 32.53$\pm$6.29\% \\
        I    & I+P  & 72.65$\pm$1.61\% \\
        P    & I    & 19.63$\pm$4.79\% \\
        P    & P    & 64.12$\pm$2.38\% \\
        P    & I+P  & 64.08$\pm$2.44\% \\
        I+P  & I    & 72.05$\pm$1.85\% \\
        I+P  & P    & 64.12$\pm$2.38\% \\
        I+P  & I+P  & 68.90$\pm$2.27\% \\
        \bottomrule
    \end{tabular}%
    } 
    \hfill
    \resizebox{.32\textwidth}{!}{%
    \begin{tabular}{C{1.5cm} C{1.5cm} C{2cm}}
        \toprule
        \multicolumn{3}{c}{ViT-Tiny + Transformer} \\
        \midrule
        Gallery & Test & Accuracy \\
        \midrule
        I    & I    & 68.48$\pm$1.44\% \\
        I    & P    & 31.14$\pm$4.47\% \\
        I    & I+P  & 69.23$\pm$1.36\% \\
        P    & I    & 21.54$\pm$3.28\% \\
        P    & P    & 64.37$\pm$2.14\% \\
        P    & I+P  & 64.36$\pm$2.13\% \\
        I+P  & I    & 68.48$\pm$1.44\% \\
        I+P  & P    & 64.37$\pm$2.14\% \\
        I+P  & I+P  & 69.14$\pm$1.75\% \\
        \bottomrule
    \end{tabular}%
    } 
    \label{tab:uto_sea_full}
\end{table}

\subsection*{A.4. Combined Training Full Results.}
\label{appendix:A4}
Full results for the combined training are presented in Tables \ref{tab:both_sea_small_full}, and \ref{tab:both_sea_large_full} for the smaller and larger model variants respectively. For the larger profile encoders we increased the base channels from $32$ to $64$ for CNN, and hidden dimension from $128$ to $256$ and feedforward dimension from $256$ to $512$ for transformer. The resulting sizes in parameters were $3.8$M and $3.7$M for CNN and transformer respectively.

% % % % % % % % % % % % % % % % % % % % % % % % 
% LAB + UTO -> SEA Small,, InfoNCE
% % % % % % % % % % % % % % % % % % % % % % % % 
\begin{table}[htb]
    \centering
    \caption{Full results, LAB + UTO $\rightarrow$ SEA, InfoNCE}
    \resizebox{.32\textwidth}{!}{%
    \begin{tabular}{C{1.5cm} C{1.5cm} C{2cm}}
        \toprule
        \multicolumn{3}{c}{EfficientNet B0 + CNN} \\
        \midrule
        Gallery & Test & Accuracy \\
        \midrule
        I    & I    & 68.46$\pm$2.94\% \\
        I    & P    & 30.21$\pm$3.37\% \\
        I    & I+P  & 68.44$\pm$3.23\% \\
        P    & I    & 37.15$\pm$4.03\% \\
        P    & P    & 64.07$\pm$2.92\% \\
        P    & I+P  & 65.11$\pm$3.03\% \\
        I+P  & I    & 68.39$\pm$2.96\% \\
        I+P  & P    & 64.07$\pm$2.92\% \\
        I+P  & I+P  & 71.05$\pm$2.87\% \\
        \bottomrule
    \end{tabular}%
    }
    \hfill
    \resizebox{.32\textwidth}{!}{%
    \begin{tabular}{C{1.5cm} C{1.5cm} C{2cm}}
        \toprule
        \multicolumn{3}{c}{ConvNeXt Femto + CNN} \\
        \midrule
        Gallery & Test & Accuracy \\
        \midrule
        I    & I    & 67.77$\pm$1.93\% \\
        I    & P    & 39.02$\pm$4.02\% \\
        I    & I+P  & 68.57$\pm$2.21\% \\
        P    & I    & 37.57$\pm$2.92\% \\
        P    & P    & 64.32$\pm$2.29\% \\
        P    & I+P  & 65.25$\pm$2.28\% \\
        I+P  & I    & 67.69$\pm$1.93\% \\
        I+P  & P    & 64.32$\pm$2.28\% \\
        I+P  & I+P  & 71.77$\pm$2.42\% \\
        \bottomrule
    \end{tabular}%
    }
    \hfill
    \resizebox{.32\textwidth}{!}{%
    \begin{tabular}{C{1.5cm} C{1.5cm} C{2cm}}
        \toprule
        \multicolumn{3}{c}{ViT-Tiny + CNN} \\
        \midrule
        Gallery & Test & Accuracy \\
        \midrule
        I    & I    & 66.23$\pm$1.84\% \\
        I    & P    & 32.42$\pm$4.50\% \\
        I    & I+P  & 66.24$\pm$2.02\% \\
        P    & I    & 32.56$\pm$4.59\% \\
        P    & P    & 65.36$\pm$2.57\% \\
        P    & I+P  & 65.92$\pm$2.86\% \\
        I+P  & I    & 66.25$\pm$1.86\% \\
        I+P  & P    & 65.34$\pm$2.57\% \\
        I+P  & I+P  & 71.37$\pm$2.17\% \\
        \bottomrule
    \end{tabular}%
    }
    \hfill
    \resizebox{.32\textwidth}{!}{%
    \begin{tabular}{C{1.5cm} C{1.5cm} C{2cm}}
        \toprule
        \multicolumn{3}{c}{EfficientNet B0 + Transformer} \\
        \midrule
        Gallery & Test & Accuracy \\
        \midrule
        I    & I    & 70.25$\pm$2.22\% \\
        I    & P    & 42.30$\pm$4.78\% \\
        I    & I+P  & 70.86$\pm$2.34\% \\
        P    & I    & 35.81$\pm$3.41\% \\
        P    & P    & 66.00$\pm$2.28\% \\
        P    & I+P  & 66.51$\pm$2.28\% \\
        I+P  & I    & 70.24$\pm$2.22\% \\
        I+P  & P    & 66.00$\pm$2.28\% \\
        I+P  & I+P  & 72.13$\pm$2.03\% \\
        \bottomrule
    \end{tabular}%
    }
    \hfill
    \resizebox{.32\textwidth}{!}{%
    \begin{tabular}{C{1.5cm} C{1.5cm} C{2cm}}
        \toprule
        \multicolumn{3}{c}{ConvNeXt Femto + Transformer} \\
        \midrule
        Gallery & Test & Accuracy \\
        \midrule
        I    & I    & 68.25$\pm$1.57\% \\
        I    & P    & 44.25$\pm$6.44\% \\
        I    & I+P  & 69.43$\pm$1.89\% \\
        P    & I    & 28.65$\pm$4.78\% \\
        P    & P    & 65.93$\pm$2.15\% \\
        P    & I+P  & 66.15$\pm$2.31\% \\
        I+P  & I    & 68.25$\pm$1.57\% \\
        I+P  & P    & 65.93$\pm$2.15\% \\
        I+P  & I+P  & 71.67$\pm$1.53\% \\
        \bottomrule
    \end{tabular}%
    } 
    \hfill
    \resizebox{.32\textwidth}{!}{%
    \begin{tabular}{C{1.5cm} C{1.5cm} C{2cm}}
        \toprule
        \multicolumn{3}{c}{ViT-Tiny + Transformer} \\
        \midrule
        Gallery & Test & Accuracy \\
        \midrule
        I    & I    & 68.86$\pm$1.39\% \\
        I    & P    & 37.49$\pm$3.54\% \\
        I    & I+P  & 70.06$\pm$1.71\% \\
        P    & I    & 33.44$\pm$1.87\% \\
        P    & P    & 66.37$\pm$1.64\% \\
        P    & I+P  & 66.97$\pm$1.68\% \\
        I+P  & I    & 68.86$\pm$1.39\% \\
        I+P  & P    & 66.37$\pm$1.64\% \\
        I+P  & I+P  & 72.73$\pm$1.20\% \\
        \bottomrule
    \end{tabular}%
    } 
    \label{tab:both_sea_small_full}
\end{table}

% % % % % % % % % % % % % % % % % % % % % % % % 
% LAB + UTO -> SEA Large,, InfoNCE
% % % % % % % % % % % % % % % % % % % % % % % % 
\begin{table}[htb]
    \centering
    \caption{Full results, LAB + UTO $\rightarrow$ SEA, InfoNCE}
    \resizebox{.32\textwidth}{!}{%
    \begin{tabular}{C{1.5cm} C{1.5cm} C{2cm}}
        \toprule
        \multicolumn{3}{c}{EfficientNet B4 + CNN} \\
        \midrule
        Gallery & Test & Accuracy \\
        \midrule
        I    & I    & 73.60$\pm$2.23\% \\
        I    & P    & 36.73$\pm$3.04\% \\
        I    & I+P  & 73.99$\pm$2.11\% \\
        P    & I    & 38.81$\pm$3.51\% \\
        P    & P    & 65.19$\pm$2.31\% \\
        P    & I+P  & 66.19$\pm$2.27\% \\
        I+P  & I    & 73.59$\pm$2.23\% \\
        I+P  & P    & 65.19$\pm$2.31\% \\
        I+P  & I+P  & 74.94$\pm$2.33\% \\
        \bottomrule
    \end{tabular}%
    }
    \hfill
    \resizebox{.32\textwidth}{!}{%
    \begin{tabular}{C{1.5cm} C{1.5cm} C{2cm}}
        \toprule
        \multicolumn{3}{c}{ConvNeXt Tiny + CNN} \\
        \midrule
        Gallery & Test & Accuracy \\
        \midrule
        I    & I    & 70.87$\pm$2.74\% \\
        I    & P    & 35.77$\pm$3.06\% \\
        I    & I+P  & 70.93$\pm$3.08\% \\
        P    & I    & 33.65$\pm$2.89\% \\
        P    & P    & 66.26$\pm$2.37\% \\
        P    & I+P  & 66.87$\pm$2.47\% \\
        I+P  & I    & 70.63$\pm$2.76\% \\
        I+P  & P    & 66.25$\pm$2.36\% \\
        I+P  & I+P  & 73.77$\pm$2.24\% \\
        \bottomrule
    \end{tabular}%
    }
    \hfill
    \resizebox{.32\textwidth}{!}{%
    \begin{tabular}{C{1.5cm} C{1.5cm} C{2cm}}
        \toprule
        \multicolumn{3}{c}{ViT-Small + CNN} \\
        \midrule
        Gallery & Test & Accuracy \\
        \midrule
        I    & I    & 72.40$\pm$2.31\% \\
        I    & P    & 41.15$\pm$4.36\% \\
        I    & I+P  & 72.59$\pm$2.54\% \\
        P    & I    & 42.56$\pm$2.51\% \\
        P    & P    & 65.78$\pm$2.42\% \\
        P    & I+P  & 66.73$\pm$2.34\% \\
        I+P  & I    & 72.19$\pm$2.35\% \\
        I+P  & P    & 65.79$\pm$2.40\% \\
        I+P  & I+P  & 72.75$\pm$2.04\% \\
        \bottomrule
    \end{tabular}%
    }
    \hfill
    \resizebox{.32\textwidth}{!}{%
    \begin{tabular}{C{1.5cm} C{1.5cm} C{2cm}}
        \toprule
        \multicolumn{3}{c}{EfficientNet B4 + Transformer} \\
        \midrule
        Gallery & Test & Accuracy \\
        \midrule
        I    & I    & 73.48$\pm$2.79\% \\
        I    & P    & 40.15$\pm$4.22\% \\
        I    & I+P  & 72.81$\pm$2.45\% \\
        P    & I    & 43.98$\pm$2.97\% \\
        P    & P    & 66.11$\pm$3.19\% \\
        P    & I+P  & 66.71$\pm$3.26\% \\
        I+P  & I    & 73.43$\pm$2.78\% \\
        I+P  & P    & 66.11$\pm$3.18\% \\
        I+P  & I+P  & 71.78$\pm$2.99\% \\
        \bottomrule
    \end{tabular}%
    }
    \hfill
    \resizebox{.32\textwidth}{!}{%
    \begin{tabular}{C{1.5cm} C{1.5cm} C{2cm}}
        \toprule
        \multicolumn{3}{c}{ConvNeXt Tiny + Transformer} \\
        \midrule
        Gallery & Test & Accuracy \\
        \midrule
        I    & I    & 68.16$\pm$1.98\% \\
        I    & P    & 44.68$\pm$3.51\% \\
        I    & I+P  & 69.00$\pm$2.05\% \\
        P    & I    & 36.60$\pm$2.65\% \\
        P    & P    & 67.27$\pm$2.85\% \\
        P    & I+P  & 67.50$\pm$3.00\% \\
        I+P  & I    & 68.08$\pm$2.00\% \\
        I+P  & P    & 67.28$\pm$2.84\% \\
        I+P  & I+P  & 71.74$\pm$2.39\% \\
        \bottomrule
    \end{tabular}%
    } 
    \hfill
    \resizebox{.32\textwidth}{!}{%
    \begin{tabular}{C{1.5cm} C{1.5cm} C{2cm}}
        \toprule
        \multicolumn{3}{c}{ViT-Small + Transformer} \\
        \midrule
        Gallery & Test & Accuracy \\
        \midrule
        I    & I    & 69.74$\pm$1.65\% \\
        I    & P    & 39.61$\pm$5.15\% \\
        I    & I+P  & 70.66$\pm$1.87\% \\
        P    & I    & 35.57$\pm$3.27\% \\
        P    & P    & 66.50$\pm$1.83\% \\
        P    & I+P  & 66.99$\pm$1.82\% \\
        I+P  & I    & 69.69$\pm$1.65\% \\
        I+P  & P    & 66.50$\pm$1.83\% \\
        I+P  & I+P  & 71.04$\pm$1.40\% \\
        \bottomrule
    \end{tabular}%
    } 
    \label{tab:both_sea_large_full}
\end{table}

\subsection*{A.5. Self-supervised Learning Baseline Configuration.}
\label{appendix:A5}
The image-only baseline DINO model was trained for $300$ epochs using the adamW optimizer with a batch size of $512$. The learning rate is linearly ramped up during the first $10$ epochs to its base value, which was set to $0.005$ for EfficientNet and $0.0001$ for ConvNeXt and ViT. The learning rate was set lower for ConvNeXt and ViT as otherwise the models did not start learning. Following the original implementation, the weight decay follows a cosine decay from $0.04$ to $0.4$. The student temperature was set to $0.1$ and the teacher temperature uses linear warm-up from $0.04$ to $0.07$ during the first $30$ epochs. 

For data augmentations, the scale bar was cropped, the remaining region padded to a square using edge padding to preserve aspect ratio, resized to $236\times236$ pixels, and converted to grayscale. Following this, we used the same augmentations that were used in the original implementation. $2$ global and $6$ local crops were used, with the local crop scale selected from ($0.1, 0.4$). For the DINO head, a hidden dimension of $256$ was used, a bottleneck dimension of $64$, and an output dimension of $512$.

\end{document}